\pdfoutput=1

\documentclass[11pt]{article}

\usepackage[preprint]{acl}

\usepackage{times}
\usepackage{latexsym}
\usepackage{listings}

\usepackage[T1]{fontenc}

\usepackage[utf8]{inputenc}

\usepackage{microtype}

\usepackage{inconsolata}

\usepackage{graphicx}

\usepackage{booktabs}
\usepackage{tabularx}
\usepackage{lipsum}  
\usepackage{caption}
\usepackage{subcaption}
\usepackage{CJKutf8}
\usepackage{xspace}
\usepackage{multirow}
\usepackage[framemethod=TikZ]{mdframed}
\usepackage{tabu}
\usepackage{graphicx}

\usepackage{listings}
\usepackage{xcolor}

\definecolor{codegreen}{rgb}{0,0.6,0}
\definecolor{codegray}{rgb}{0.5,0.5,0.5}
\definecolor{codepurple}{rgb}{0.58,0,0.82}
\definecolor{backcolour}{rgb}{0.95,0.95,0.92}

\lstdefinestyle{mystyle}{
    backgroundcolor=\color{backcolour},   
    commentstyle=\color{codegreen},
    keywordstyle=\color{magenta},
    numberstyle=\tiny\color{codegray},
    stringstyle=\color{codepurple},
    basicstyle=\ttfamily\footnotesize,
    breakatwhitespace=false,         
    breaklines=true,                 
    captionpos=b,                    
    keepspaces=true,                 
    numbers=left,                    
    numbersep=5pt,                  
    showspaces=false,                
    showstringspaces=false,
    showtabs=false,                  
    tabsize=2
}

\usepackage{amsmath}
\usepackage{amssymb}
\usepackage{mathtools}
\usepackage{amsthm}
\usepackage{amsmath}
\usepackage{blkarray}
\usepackage{cleveref}

\usepackage[subtle]{savetrees}

\definecolor{cb-0}{RGB}{216, 27, 96}
\definecolor{cb-1}{RGB}{30,136,229}
\definecolor{cb-2}{RGB}{255,193,7}
\definecolor{cb-3}{RGB}{0, 77, 64}
\definecolor{cb-4}{RGB}{150,220,174}

\NewDocumentCommand{\david}{ mO{} }{\textcolor{red}{\textsuperscript{\textit{David}}\textsf{\textbf{\small[#1]}}}\PackageError{TODO}{You have unresolved todos}{(You should probably do this!)}}
\NewDocumentCommand{\jiaxin}{ mO{} }{\textcolor{blue}{\textsuperscript{\textit{Jiaxin}}\textsf{\textbf{\small[#1]}}}\PackageError{TODO}{You have unresolved todos}{(You should probably do this!)}}  
\NewDocumentCommand{\rudy}{ mO{} }{\textcolor{green}{\textsuperscript{\textit{Rudy}}\textsf{\textbf{\small[#1]}}}\PackageError{TODO}{You have unresolved todos}{(You should probably do this!)}}  
\NewDocumentCommand{\ritwik}{ mO{} }{\textcolor{purple}{\textsuperscript{\textit{Ritwik}}\textsf{\textbf{\small[#1]}}}\PackageError{TODO}{You have unresolved todos}{(You should probably do this!)}}  
\NewDocumentCommand{\eric}{ mO{} }{\textcolor{orange}{\textsuperscript{\textit{Eric}}\textsf{\textbf{\small[#1]}}}\PackageError{TODO}{You have unresolved todos}{(You should probably do this!)}}

\title{Enough Coin Flips Can Make LLMs Act Bayesian}

\author{Ritwik Gupta\thanks{Denotes co-first authorship.}~\quad
Rodolfo Corona\footnotemark[1]~\quad
Jiaxin Ge\footnotemark[1]~\quad
Eric Wang~\quad~\\
\textbf{Dan Klein}~\quad
\textbf{Trevor Darrell}~\quad
\textbf{David M. Chan} \\[8pt]
University of California, Berkeley
}

\begin{document}
\maketitle

\begin{abstract}
Large language models (LLMs) exhibit the ability to generalize given few-shot examples in their input prompt, an emergent capability known as in-context learning (ICL). We investigate whether LLMs use ICL to perform structured reasoning in ways that are consistent with a Bayesian framework or rely on pattern matching. Using a controlled setting of biased coin flips, we find that: (1) LLMs often possess biased priors, causing initial divergence in zero-shot settings, (2) in-context evidence outweighs explicit bias instructions, (3) LLMs broadly follow Bayesian posterior updates, with deviations primarily due to miscalibrated priors rather than flawed updates, and (4) attention magnitude has negligible effect on Bayesian inference. With sufficient demonstrations of biased coin flips via ICL, LLMs update their priors in a Bayesian manner. Code and visualizations are available on the \href{https://ai-climate.berkeley.edu/llm-coin-flips/}{project page}.
\end{abstract}



\section{Introduction}

Large language models (LLMs) designed for next-token prediction have gained significant popularity, largely because of their ability to generalize beyond language prediction, and perform a wide range of novel tasks without requiring explicit weight updates \cite{brown2020language}. Methods to induce emergence in controlled ways include techniques such as chain-of-thought prompting \cite{wei2022chain}, prompt chaining \cite{wu2022ai}, and in-context learning (ICL). ICL, particularly, provides demonstrations of a specific task to the model as part of its input prompt.

\begin{figure}

    \small

    \centering

    \includegraphics[width=\linewidth]{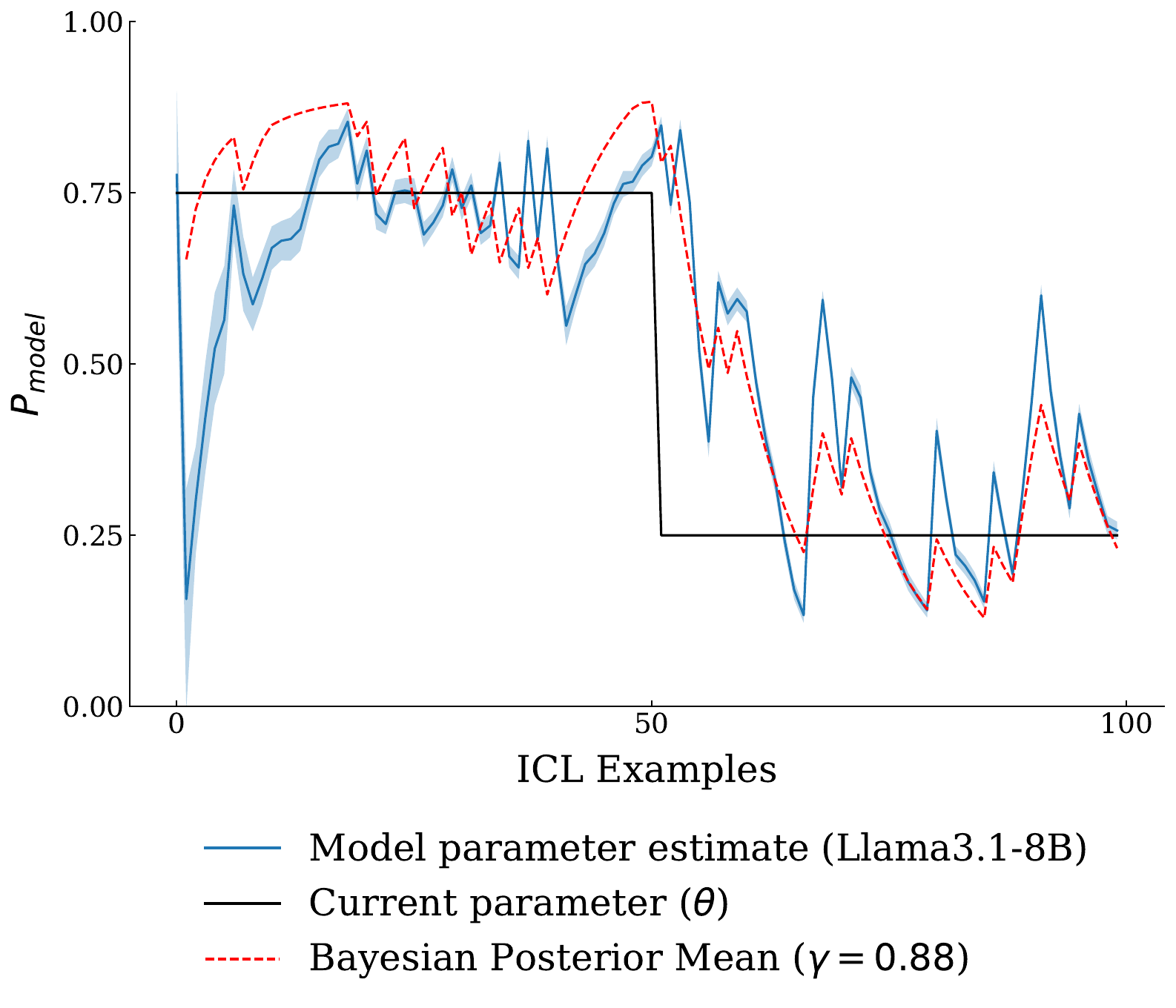}

    \caption{When we ask large language models (LLMs) to model sequences with in-context learning (ICL), how do they adapt their posterior probabilities given the provided examples? This figure explores how model probabilities change as we add new ICL examples in a biased coin-flipping experiment. The X-axis represents steps in the trajectory, while the Y-axis shows the predicted parameter of a Bernoulli distribution. Our results reveal that, while LLMs often have poorly calibrated priors, their updated parameter estimates broadly align with Bayesian behavior. }

    \label{fig:teaser}

\end{figure}

Despite significant empirical success, the underlying mechanisms of ICL remain poorly understood. While it is clear that models can adapt their predictions in response to few-shot examples, it is less clear whether this adaptation aligns with statistical principles such as Bayesian inference. Do these models simply replicate memorized patterns from their training data, or do they systematically update their beliefs in a way that is consistent with Bayesian reasoning when presented with new evidence in the prompt? In this work, we investigate these questions using a controlled setting of biased coin flips.

A prominent explanation for ICL’s behavior is that it reflects some form of Bayesian learning. Prior studies have suggested that, in certain scenarios, large language models can approximate Bayesian updating by maintaining an implicit prior distribution over latent structures and refining that prior using contextual information \cite{xie2021explanation,hahn2023theory,akyurek2022learning,zhang2023and,panwar2023context}. However, many of these works rely on tasks (e.g., question-answering or language modeling) where the true posterior distribution is unknown, making it difficult to determine how closely a model’s inferences adhere to normative Bayesian updates. Other research has pursued more controlled theoretical settings with known posteriors, but with strong assumptions about model architectures or data domains. As a result, the extent to which pre-trained LLMs truly follow Bayesian update rules, and whether their test-time behavior aligns with canonical probabilistic reasoning, remains an open question.

We reduce the complexity of typical ICL analyses by focusing on a stochastic phenomenon: biased coin flips. This setting allows us to compute all relevant Bayesian quantities and thus precisely evaluate whether pre-trained LLMs update their priors in a Bayesian manner. By examining how models estimate coin biases and incorporate sequential evidence, we can directly assess the degree to which they converge on normative probabilistic reasoning. In addition, this streamlined setup lets us explore the impact of factors like attention, model scale, and instruction tuning without introducing the distributional complexities of more elaborate language tasks.

In this work we find several results: (1) language models often exhibit biased priors for stochastic phenomena, leading to significant initial divergence when modeling zero-shot scenarios; (2) they tend to disregard explicit bias instructions and rely more heavily on in-context examples; (3) their predictions are consistent with Bayesian updates once new evidence is presented, with most deviations from the true posterior arising from miscalibrated priors rather than faulty updates; and (4) attention magnitude has minimal influence on the updating process. Taken together, these results imply that LLMs implicitly perform Bayesian modeling in simple cases, and that poor priors may cause reduced performance in more complex environments rather than failures of updates due to in-context learning.
\section{Background \& Related Work}

\paragraph{Representing probabilities in language models.} As LLMs have proliferated across a wide set of applications, many have examined whether LLMs can properly represent the concept of probability. Much of this examination has been done through the lens of model calibration and alignment. \citet{zhuIncoherentProbabilityJudgments2024} show that LLMs are biased judges of probability much in the same fashion as human probability judgments. \citet{guLLMsPlayDice2024} asks whether LLMs can play dice and finds that while LLMs know what probability is, they struggle to accurately sample from distributions. They attempt to solve this through tool use, but find that this is not a guaranteed solution to the problem. \citet{meisterBenchmarkingDistributionalAlignment2024} evaluates how well LLMs can align to human groups’ distributions over a diverse set of opinions. They find that LLMs are good at describing biased distributions but are incapable of simulating these distributions.

In this work, we explore the ability of LLMs to simulate biased probability distributions and explore the mechanism of in-context learning as a natural method by which LLMs can align their priors to requested distributions.

\paragraph{In-context learning.} \citet{brown2020language} introduces in-context learning (ICL) as a mechanism for few-shot generalization in language models. Although ICL usage has surged, users rarely employ it as a method to align models with target distributions. Further, issues with models’ sensitivity to the positioning of tokens in their prompts have complicated the effective use of ICL as an alignment technique. \citet{luFantasticallyOrderedPrompts2022} demonstrates that the positioning of information within an ICL prompt affects model performance and devises a permutation-based approach to overcome this bias. \citet{liuLostMiddleHow2023} extends this analysis to highlight a persistent ``lost-in-the-middle’’ effect, in which there is implicit positional bias for information as it relates to accuracy and suggest that the mechanism behind the lost-in-the-middle effect may be more closely related to position embedding. \citet{liu2024llms} show that LLMs can extrapolate the behavior of dynamical systems given large numbers of in-context examples. However, their discovered power-law fits imperfectly, demonstrating high loss at long contexts.

Our work explores a time-varying discount factor for in-context learning, more directly explaining the higher-than-expected loss at long context lengths. We demonstrate that in-context rollouts of a probability distribution correlate well with the mean of a Bayesian posterior. Further, we analyze how attention weights affect output accuracies and find little correlation.

\paragraph{Bayesian updating in language models.} Many authors have explored the mechanisms through which ICL capability emerges in language models. \citet{xie2021explanation} finds that ICL can be viewed as a language model implicitly performing Bayesian inference---i.e., ICL emerges via modeling long-range coherence during pretraining. \citet{jiangLatentSpaceTheory2023} shows that emergent capabilities of LLMs, such as ICL, are Bayesian inference on the sparse joint distribution of languages. \citet{wangLargeLanguageModels2024} react to the ordering sensitivity of ICL prompts and pose ICL as a natural side effect of LLMs functioning as latent variable models. Finally, \citet{zhang2023and} posit that ICL is an implicit form of Bayesian model averaging.

A complementary perspective comes \citet{zhao21c}. They demonstrate that a model’s outputs in few-shot prompts can be systematically skewed by inherent biases or the arrangement of examples. They show that adjusting the model’s decision boundary or distribution (via contextual calibration) can substantially mitigate these biases.

Our own findings, that LLMs can often apply Bayesian-like updates despite relying on miscalibrated priors, resonate with this need for calibration, underscoring the importance of correcting initial biases when using LLMs in downstream tasks. We confirm the ordering sensitivity of ICL prompts and further show empirically that ICL has several implicit Bayesian modeling behaviors. Finally, we demonstrate that it is unlikely that attention magnitude is a key component of this formalization.

\section{Preliminaries}

\paragraph{Bayesian systems:} General Bayesian systems are expected to update their beliefs in a manner consistent with Bayes' rule. Given some evidence, D, a prior distribution $p(\theta)$ and a likelihood $p(D|\theta)$, the posterior distribution is obtained via:
\begin{equation}
    p(\theta|D) = \frac{p(D|\theta) p(\theta)}{p(D)}
\end{equation}
where $p(D)$ is the marginal likelihood (or evidence) ensuring the posterior is properly normalized. While prior work \cite{falck2024context} has explored additional assumptions (such as exchangeability), here we aim to explore the fundamental update process in a restricted environment.

\paragraph{Modeling coin-flips as Bayesian processes:} In our setup, we model a biased coin by treating the probability of obtaining heads, denoted by $\theta$, as a random variable with a binomial distribution. Suppose we perform $n$ independent coin flips and observe $k$ heads and $n-k$ tails. The likelihood of the observed data is given by:
\begin{equation}
p(D|\theta) = \theta^k (1-\theta)^{n-k}
\end{equation}

A common choice for the prior distribution of $\theta$ is the Beta distribution due to its conjugacy with the binomial likelihood:
\begin{equation}
p(\theta) = \frac{\theta^{\alpha-1}(1-\theta)^{\beta-1}}{B(\alpha,\beta)}
\end{equation}
where $B(\alpha,\beta)$ is the Beta function. By applying Bayes' theorem, the posterior distribution is thus proportional to the product of the likelihood and the prior:
\begin{align}
    p(\theta|D) &\propto p(D|\theta) p(\theta) \\
     & \propto \theta^k (1-\theta)^{n-k} \cdot \theta^{\alpha-1}(1-\theta)^{\beta-1} \\
    & = \theta^{\alpha+k-1}(1-\theta)^{\beta+n-k-1}
\end{align}
And the posterior distribution for $\theta$ is also a Beta distribution:
\begin{equation}
\theta|D \sim \text{Beta}(\alpha+k, \beta+n-k).
\end{equation}

It is often useful to consider the case where we have no strong prior beliefs about the coin's bias, leading us to adopt a uniform prior for $\theta$. The uniform prior over the interval \([0,1]\) is a special case of the Beta distribution with parameters \(\alpha = 1\) and \(\beta = 1\), i.e., $p(\theta) = \text{Beta}(\theta; 1,1) = 1$. When using the uniform prior, the posterior distribution becomes:
\begin{equation}
p(\theta|D) \propto \theta^k (1-\theta)^{n-k},
\end{equation}

This Bayesian framework allows us to update our beliefs about the coin's bias as more coin-flip data is collected, providing both a point estimate and a measure of uncertainty for $\theta$.

\paragraph{Experimental design:} We focus on open-source language models and extract stochastic representations directly from the underlying learned model distributions. Consider a sequence of tokens
\begin{equation}
    x = \{x_1,x_2,\dots,x_n\}
\end{equation}
drawn from a vocabulary $V$ (with \(|V|\) elements). A large next-token prediction-based language model, \(\mathcal{M}\), approximates a probability distribution over the next token:
\begin{equation}
    p_{\mathcal{M}}(x_{i+1}\mid x_{1:i})
\end{equation}
where $x_{1:i} = \{x_1, x_2, \dots, x_i\}$.

To evaluate stochastic processes, we define a fixed set of possible outcomes \(\Omega = \{o_1, o_2, \dots, o_k\}\), where each outcome $o \in \Omega$ is a sequence of tokens corresponding to a specific string value (e.g., when modeling a coin flip, the outcomes “heads” and “tails” might correspond to token sequences \verb|[_heads]| and \verb|[_tails]|, respectively). For each outcome $o$, we compute the probability given a prompt—analogous to updating our beliefs in a Bayesian framework—as follows:
\begin{equation}
    p_{\mathcal{M}}(o\mid \text{prompt}) = \prod_{i=1}^{|o|} p_{\mathcal{M}}(o_i\mid o_{1:i-1},\text{prompt})
\end{equation}
where \(|o|\) denotes the number of tokens in $o$ and $o_{1:i-1}$ represents the subsequence of tokens preceding the $i$th token in $o$.

Because these outcomes are a subset of all possible token sequences that \(\mathcal{M}\) could generate, we renormalize the distribution over the support \(\Omega\). We denote the renormalized model distribution as \(\hat{p}_{\mathcal{M}}(o)\) for $o \in \Omega$ (see \autoref{app:methods_softmax} for further details on the renormalization process).

In our experiments, we measure the total variation distance (TVD) between the true posterior distribution $p^*(o)$ and the normalized model distribution \(\hat{p}_{\mathcal{M}}(o)\) over the support \(\Omega\):
\begin{equation}
    \delta(p^*,\hat{p}_{\mathcal{M}}) = \frac{1}{2}\sum_{o \in \Omega} \left|p^*(o) - \hat{p}_{\mathcal{M}}(o)\right|
\end{equation}
This distance metric quantifies the discrepancy between the two distributions—zero indicating perfect alignment and higher values indicating greater divergence.

We would like to clearly state that we are not claiming that LLMs themselves are explicitly Bayesian, rather, we ask the question: \textit{do model predictive distributions have Bayesian behavior?} In this paper we treat models themselves as point-wise estimators of distributional parameters (in our case, we use them to estimate the parameters of a binomial distribution), and ask if those point-wise estimates align with reasonable Bayesian frameworks. 

We evaluate several models, including Gemma-2 \cite{team2024gemma}, Phi-2/Phi-3.5 (mini) \cite{abdin2024phi}, Llama-3.1 (8B) \cite{dubey2024llama}, Mistral 7B \cite{jiang2023mistral}, and OLMoE (7B) \cite{muennighoff2024olmoe}, along with their instruction-tuned variants. For scaling experiments, we leverage the Pythia Scaling Suite \cite{biderman2023pythia}  For more details regarding these models, please refer to \autoref{app:models}.

\section{Understanding the LLM Prior}
\label{sec:prior}

Due to data-intensive pre-training, language models inherently encode a prior over $\theta$ (the likelihood of heads in the coin-flip). We are interested in understanding these priors and understanding how to update the priors via explicit prompting.

\begin{figure}
    \centering
    \includegraphics[width=\linewidth]{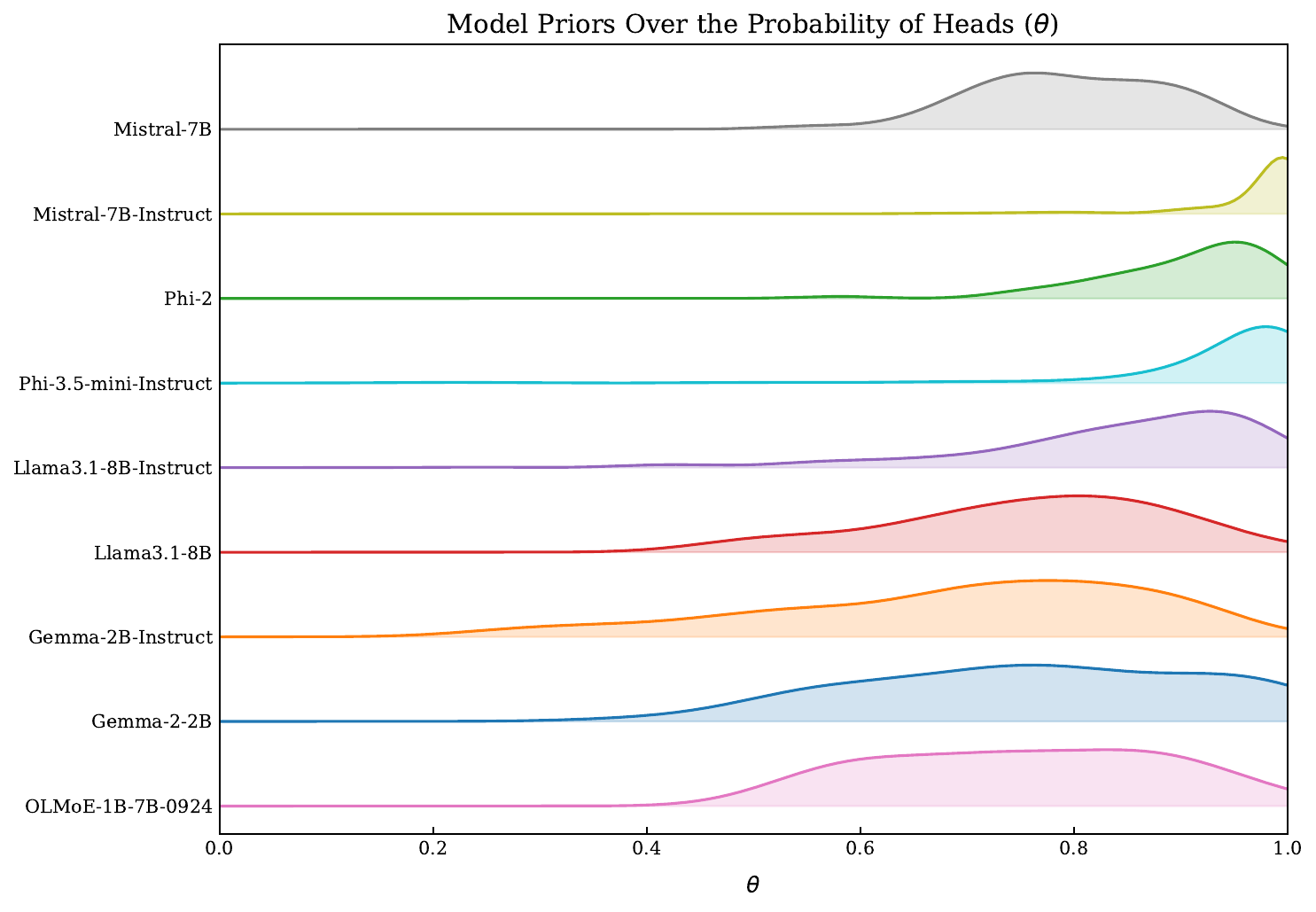}
    \caption{\textbf{Model priors:} All language models evaluated present a bias towards heads.}
    \label{fig:model_heads_priors}
\end{figure}

To extract a prior over heads and tails, we query the models for a coin flip through $50$ different prompt variants (e.g. \texttt{``I flipped a coin and it landed on'')}, and compute the normalized logit value ascribed to heads (discussed in detail in \autoref{app:methods}). As shown in \autoref{fig:model_heads_priors}, all language models evaluated begin with fundamental priors for $\theta$ that are heads-biased, and in some cases, significantly so. This observation is reflected in the tokenization structure itself; in some cases, models do not see sufficient data to assign a full token to \texttt{[\_tails]} and instead encode this in a pair of tokens (which we handle when computing probability, see \autoref{app:methods}). Thus, models begin divergent from an unbiased estimate of coin priors. 

\paragraph{Effect of explicit biasing via prompting.} Next, we explore if we can encourage models to update their priors by providing an explicit value for $\theta$ in the prompt. 
We define a set of biasing statements, i.e. describing unfair coins, of the form \texttt{``When I flip coins, they land on heads X\% of the time.''}, and run a set of trials, evaluating the TVD between models' probabilities over outcomes and the expected distribution for the biased $\theta$.

\begin{figure*}
    \centering
    \hspace{\stretch{1}}
    \includegraphics[width=0.49\linewidth]{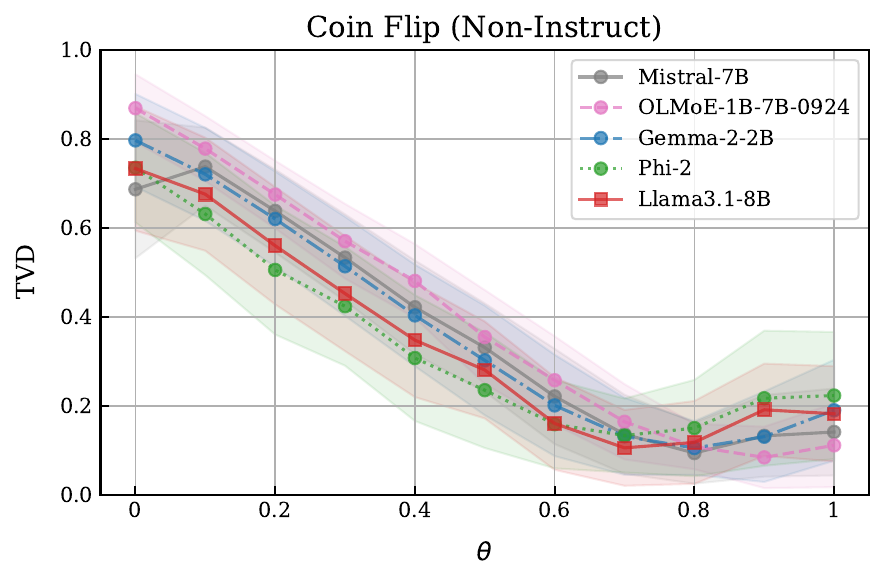}
    \hspace{\stretch{1}}
    \includegraphics[width=0.49\linewidth]{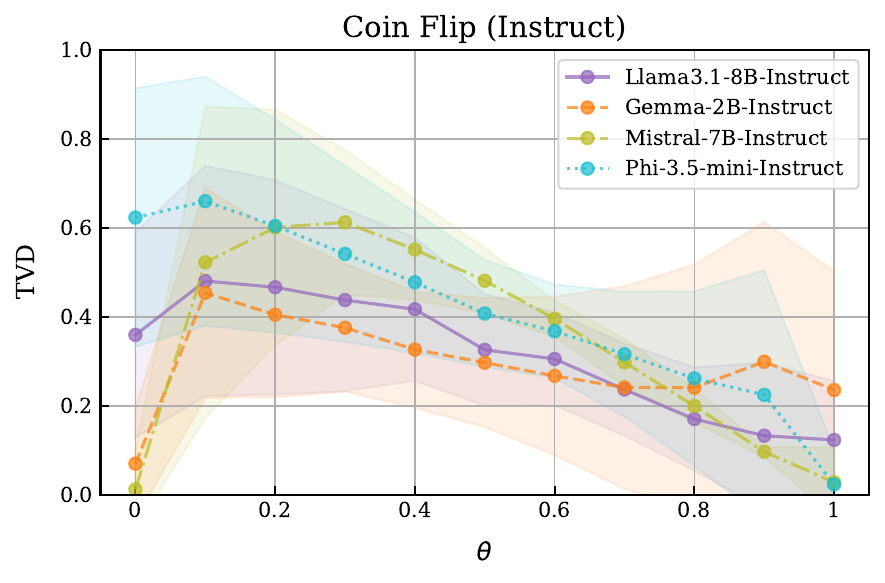}
    \hspace{\stretch{1}}
    \caption{\textbf{Biased coins:} Plots of mean total variation distance (TVD, $\downarrow$) against bias ($\theta$) for non-instruct (left) and instruct (right) models when aggregated across prompts (N=50) for the biased coin flip experiment. Shaded areas show one standard deviation. While non-instruct models both (1) ignore biasing instructions in the prompts and (2) almost always generate a biased distribution ($\approx 70\%$ heads), instruct-based models pay better attention to biasing information, and perform significantly better when modeling extreme bias (always generating heads/tails).}
    \label{fig:coin_flip}
\end{figure*}

Results from this experiment are presented in \autoref{fig:coin_flip}. 
Given an explicit bias in the input prompt, non-instruct LLMs fail to converge to the expected biased distribution with their token probabilities following their originally computed prior---generally showing a tendency to ascribe $\approx$ 60\%-80\% probability to heads, independent of explicit context. 
Instruct models performed slightly better, though they still exhibited a bias toward heads.
Additionally, instruct models showed improved performance at the extremes of bias values, with TVD values dropping for 0\% and 100\% heads biases (matching observations from \citet{zhaoCalibrateUseImproving2021}).

\begin{figure}[htbp]
    \centering
    \includegraphics[width=\linewidth]{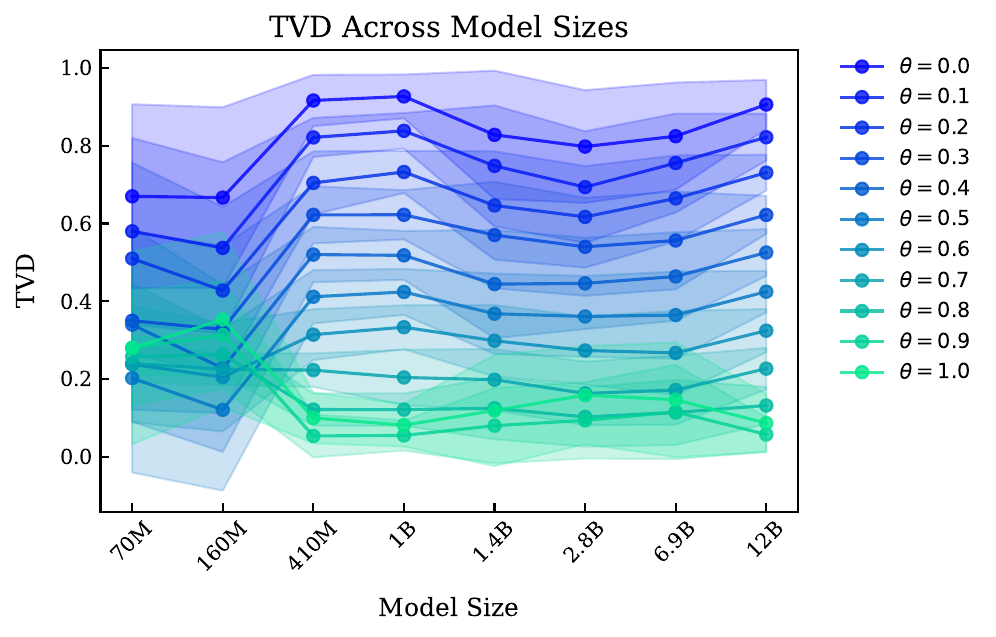}
    \caption{\textbf{Biased coins and parameter scaling:} Mean total variation distance (TVD, $\downarrow$) vs. model size for different bias percentages. We use the models from the Pythia Scaling Suite. As the size of the model increases, the performance does not change for a certain bias. The relative ordering among different biases does shift as the model size increases}
    \label{fig:scale_prior}
\end{figure}

\paragraph{Effect of model size on priors.}
Scaling the language model size has shown effectiveness in many tasks. 
Therefore, we explore whether scaling also boosts performance on modeling expected biased distribution.
We use the Pythia Scaling Suite~\cite{biderman2023pythia}, which covers model sizes ranging from 70M to 12B, and test on different biased $\theta$.
Results from this experiment are presented in \autoref{fig:scale_prior}. For a given bias, scaling the model size does not substantially change the language models' priors or improve the performance of modeling expected distributions. However, the relative ordering among different biases does shift as the model size increases.

\begin{figure*}
    \centering
    \includegraphics[width=0.49\linewidth]{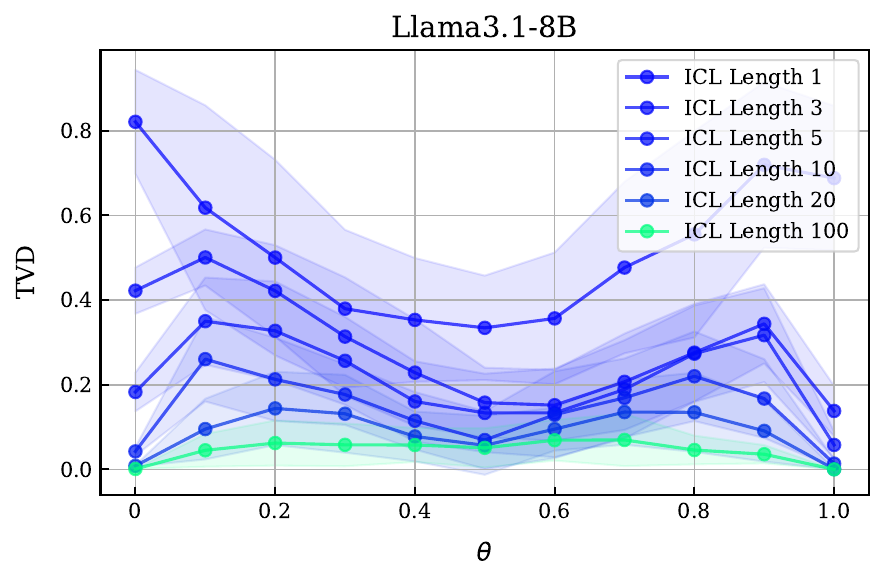}
    \includegraphics[width=0.49\linewidth]{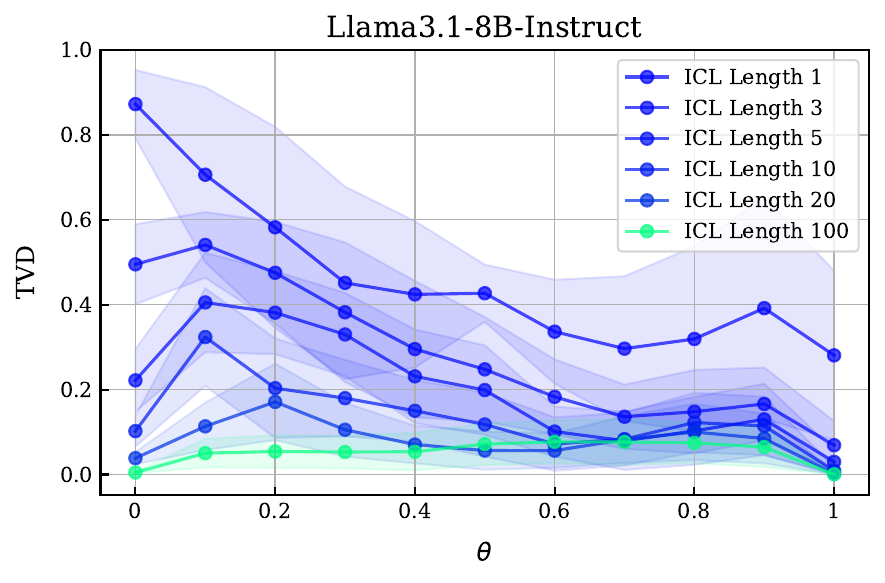}
    
    \caption{\textbf{Biased coins and ICL:} Mean total variation distance (TVD, $\downarrow$) vs. bias percentage for several ICL example lengths for Llama3.1-8B model (left) and Llama3.1-8B-Instruct (right). As the number of in-context samples increases, the performance of the models at modeling the stochastic process improves as well. Notably, adding as few as 3 in-context examples significantly improves performance, but even adding 100 in-context examples does not fully allow the model to capture the biased distribution. For other models, see \autoref{app:multi_coin_flip}. }
    \label{fig:coin_flip_icl}
\end{figure*}

\section{Does In-Context Learning Improve Parameter Estimates?} 
\label{sec:posterior}

We are interested in understanding if and how LLMs incorporate in-context evidence into their posteriors.
Specifically, rather than explicitly describing the underlying distribution as before, we implicitly specify it by providing the LLM with a sequence of samples from that distribution in its prompt (e.g., \texttt{``I flipped a coin and it landed on heads, then on tails, then on tails, then on tails, then on...''} for a coin biased toward tails). We then assess the expected distribution of the coin flip outcomes under each model after presenting these ICL prompts. 

\autoref{fig:coin_flip_icl}, shows results from the coin flip experiment on Llama-3.1-8B and Llama-3.1-8B-Instruct (see \autoref{app:multi_coin_flip} for results from other models). 
We find that models converge to the expected distribution as more evidence is provided via in-context learning.

\subsection{Effect of model scale}\label{sec:icl_scaling}

\begin{figure}
    \centering
    \includegraphics[width=\linewidth]{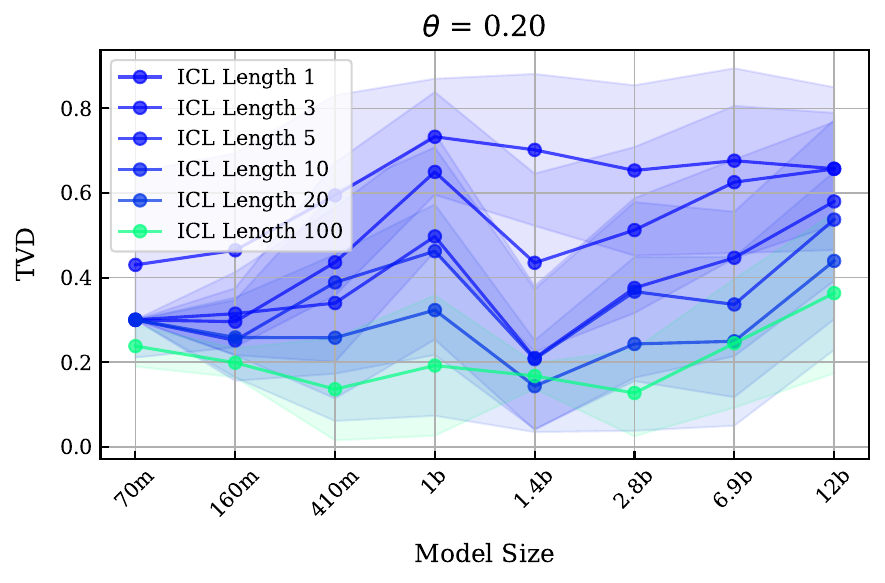}
    \caption{\textbf{ICL and parameter scaling:} Mean total variation distance (TVD, $\downarrow$) vs. model size across the Pythia Scaling Suite family with a biasing statement for $\theta = 0.20$. Model size does not have a clear impact on the benefits from ICL.}
    \label{fig:icl_scaling}
\end{figure}

 We investigate if larger models are better able to incorporate in-context-based evidence. \textit{Chinchilla}-scaling \citet{hoffmannTrainingComputeOptimalLarge2022} would suggest that larger models would also have more powerful emergent behaviors such as ICL.

 In \autoref{fig:icl_scaling}, we show the results of running the ICL experiments on the Pythia Suite for $\theta = 0.20$ (See \autoref{app:icl_scaling} for all settings of $\theta$). 
 Although ICL performance generally improves as the number of examples grows, we find that model scale has negligible impact on order dynamics, with models performing comparably across scales. Surprisingly, however, larger models appear worse at incorporating model updates on the whole, with most TVD values higher for the 12B model compared to their respective smaller models. 

\subsection{Do models perform pure Bayesian updates?}
\label{sec:pure}

To explore if models actually perform Bayesian updates during a single trial, we look directly at several ``online'' ICL trajectories. To generate these trajectories, instead of drawing trajectories entirely from a single distribution, we instead model a generative process containing $100$ steps, where the first $50$ samples are drawn $\sim Bernoulli(\theta_1)$ and the second $50$ samples are drawn $\sim Bernoulli(\theta_2)$, where $\theta_1=0.75$ and $\theta_2=0.25$. This trajectory, shown in \autoref{fig:teaser} (the black line), gives a moving target which evolves over time for the model to approximate. In this dynamic environment, we then explore how well the LLM's pointwise estimates are modeled by a Bayesian update process.

\begin{figure*}
    \centering
    \small
    \includegraphics[width=\linewidth]{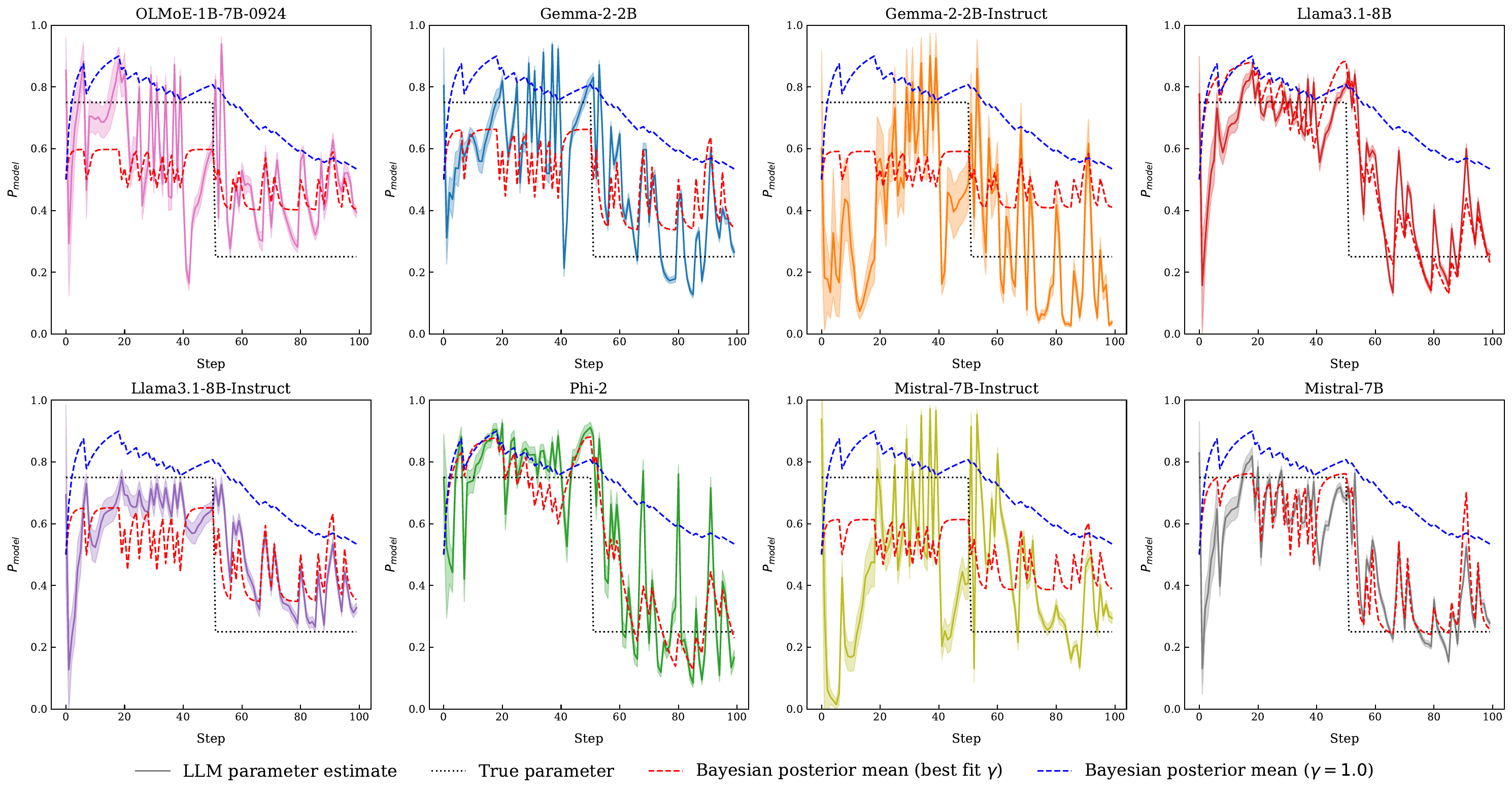}
    \caption{\textbf{Posterior evolution during Bayesian filtering:} The figure shows a single rollout of classical Bayesian filtering alongside model predictive probabilities in a 100-sample coin flip ICL task. While the overall shape of the model’s predictions aligns with Bayesian updates, the direct application of standard Bayesian filtering ($\gamma=1.0$) does not fully explain the observed behavior. Instead, the empirical fit suggests that models implicitly apply a localized Bayesian update with a shorter time horizon, aligning better with a slightly discounted filtering process. }
    \label{fig:posterior_chain}
\end{figure*}

To define this Bayesian update process, we first note that classical Bayesian filtering updates a Beta prior $\text{Beta}(\alpha, \beta)$ with each observation, treating all data equally. Given a prior and a binomial likelihood, the posterior is also Beta-distributed:
\begin{equation}
p(\theta | D) = \text{Beta}(\alpha + k, \beta + n - k),
\end{equation}
where $k$ is the number of heads observed in $n$ coin flips. 

In dynamic environments, on the other hand, recent data may be more relevant. To model this, we can introduce an exponential decay factor $\gamma$, modifying the updates to:
\begin{align}
\alpha &\leftarrow \gamma \alpha + I(H), \quad \beta \leftarrow \gamma \beta + I(T)
\end{align}
where $I(H)$ and $I(T)$ indicate the latest result. This ensures older observations gradually contribute less, allowing the model to adapt. The posterior mean remains:
\begin{equation}
\mathbb{E}[p] = \frac{\alpha}{\alpha + \beta}
\end{equation}
This decay ensures older data contributes less, allowing adaptation to shifts in $\theta$.  For $\gamma = 1.0$, this remains the classical Bayesian filtering update.

Returning to our environment, \autoref{fig:posterior_chain} shows a single example roll-out of both classical and the gamma-modified Bayesian filter, along with the associated model probabilities. We can see that while the general shape of the trajectory fits the model behavior, pure Bayesian filtering (i.e. $\gamma=1.0$) alone does not explain the behavior of the model. Instead, using a $\gamma < 1$, implying a shortened time horizon, fits the behavior almost perfectly in some cases, empirically suggesting that models are performing local Bayesian updates with a slight discount factor. 

Extending this idea, we leverage L-BFGS-B \citet{zhuAlgorithm778LBFGSB1997} to fit a $\gamma$ value to each model, with the results shown in \autoref{tab:posterior_fit_gammas}. We can see in this table that the value of $\gamma$ is notably different for each model, suggesting that models have architecture-specific time-horizon behavior. Interestingly, instruction-tuned models generally have much lower $\gamma$ values than their non-instruction-tuned counterparts. This implies that these models may be more local when performing ICL and are more willing to switch behaviors when prompted with new ICL evidence. 

\begin{table}
    \small
    \centering
    \caption{Bayesian filtering best fit $\gamma$ value. }
    \label{tab:posterior_fit_gammas}
    \begin{tabularx}{\linewidth}{Xc}
        \toprule
        \textbf{Model} & \textbf{Best-Fit $\gamma$} \\
        \midrule
        OLMoE-1B-7B-0924 & 0.3268 \\
        Gemma-2-2B & 0.4910 \\
        Gemma-2-2B-Instruct & 0.3087 \\
        Llama3.1-8B & 0.8807 \\
        Llama3.1-8B-Instruct & 0.4655 \\
        Phi-2 & 0.8781 \\
        Mistral-7B & 0.6903 \\
        Mistral-7B-Instruct & 0.9107 \\
        \bottomrule
    \end{tabularx}
\end{table}

\subsection{Does attention impact updates?}

Some prior work, such as \citet{zhang2023and}, suggests that attention helps to weight the Bayesian update. In this section, we aim to leverage our simplified setup to empirically understand the impact that attention has on the convergence behavior of the model. We use the same setup as \Cref{sec:pure} with a sequence $L$ of length $N=100$. There is a ``switchover'' point $K=50$ such that samples $L_{1-K} \sim \operatorname{Binom}(K, \theta_1)$ and $L_{K-N} \sim \operatorname{Binom}(N-K, \theta_2)$. We experiment varying $K \in [10, 90]$.

\begin{figure}
    \centering
    \includegraphics[width=\linewidth]{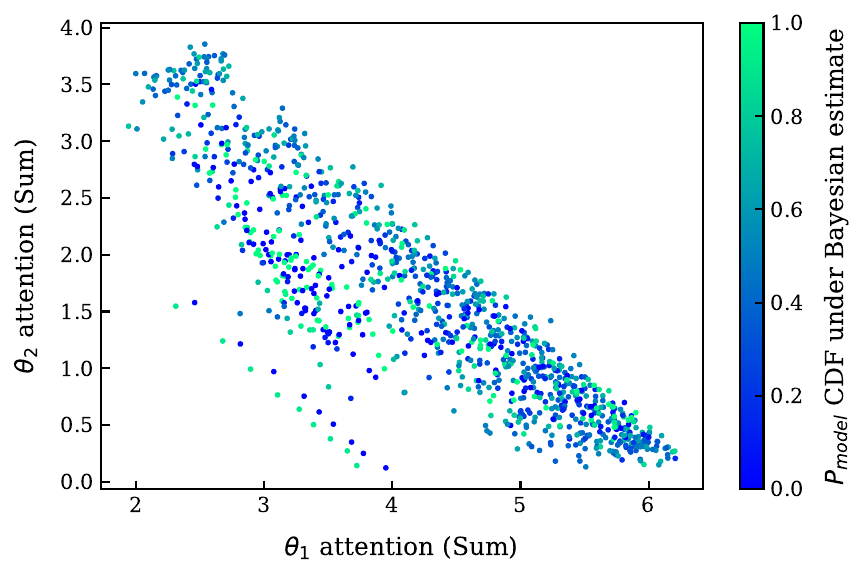}
    \caption{Relationship between total attention and model point-estimate extremity under the Bayesian posterior ($\gamma=1.0$), and all values of $K$. Overall, the extremity of the model point estimate under the Bayesian model appears uncorrelated with the attention. }
    \label{fig:attention}
\end{figure}

\autoref{fig:attention} plots the relationship between total attention and model point-estimate extremity under the Bayesian posterior ($\gamma=1.0$) (i.e. the value of the CDF of the true posterior at the model point estimate) for all $K$. We can see that the amount of attention paid to any segment is generally uncorrelated with the overall quality of the point estimate ($\theta_1: (R=0.02, p=0.48),\: \theta_2:(R=-0.03,p=0.36)$), suggesting that the total magnitude of the attention paid to each segment does not dramatically impact model quality.

\begin{figure}
    \small
    \centering
    \includegraphics[width=\linewidth]{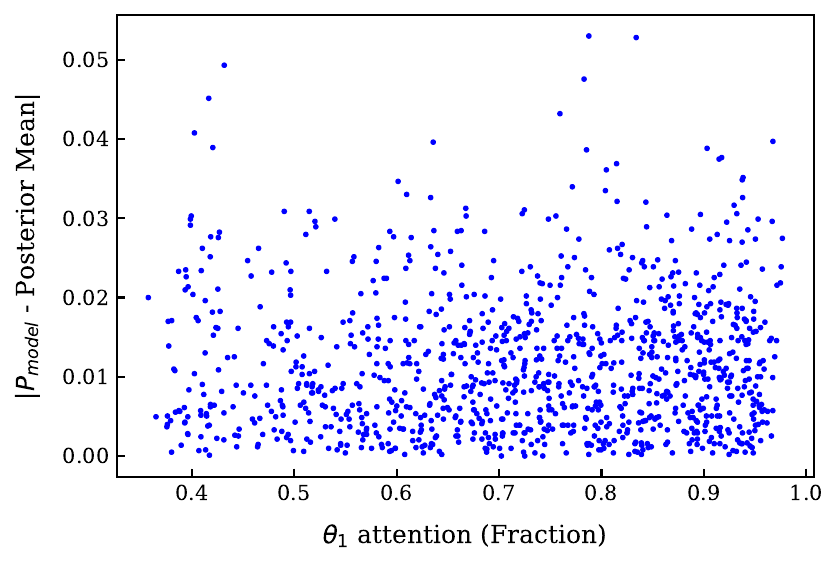}
    \caption{Fraction of attention assigned to samples from $\theta_1$ versus the deviation between the model-predicted distribution and the true posterior mean for Llama-3.1-8B for all values of $K$. The findings suggest that the relative attention paid to in-context examples does not directly predict the model's update performance.}
    \label{fig:att_frac}
\end{figure}

In addition, the fraction of attention, for all $K$, has a similar lack of correlation, as shown in \autoref{fig:att_frac}, which suggests that paying any special attention (in terms of magnitude) to any particular ICL example is uncorrelated with downstream performance during model updates. Additionally, there is no significant difference in results as we vary $K$, visualized in \autoref{app:attentional_correlation}.

\begin{figure}
    \small
    \centering
    \includegraphics[width=\linewidth]{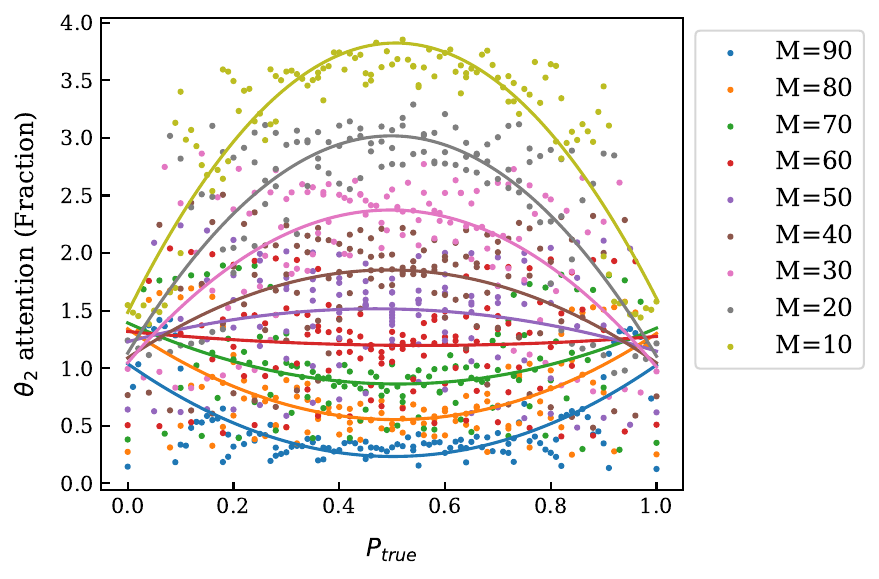}
    \caption{The fraction of attention on samples from $\theta_2$ vs. the true posterior distribution of the mixture for different values of M for Llama-3.1-8B. Lines represent the degree-2 line of best fit. When $M$ is low, the model primarily attends to $\theta_2$ when it aligns with $\theta_1$. As $M$ increases, the model pays more attention to $\theta_2$ when it significantly influences the final distribution.}
    \label{fig:att_true}
\end{figure}

Interestingly, an important indicator of attention is the (non-estimated) true parameter value. We can see in \autoref{fig:att_true} that when M is low (i.e. few samples are drawn from $\theta_2$, the model only pays attention to $\theta_2$ when it matches the $\theta_1$ distribution. When $M$ is high, the model pays attention more to samples from $\theta_2$ when $\theta_2$ is more likely to bias the distribution. These observations support a nuanced view of model attention: models pay relatively more attention to data which is more likely to lead to changes in the final distribution, but higher/lower attention is somewhat uncorrelated with final model quality. 

\section{Discussion \& Conclusion}

Our study investigated how large language models (LLMs) adapt to simple, controlled stochastic processes—specifically biased coin flips—when performing in-context learning (ICL). By stripping away complexities found in prior ICL studies, we isolated how pre-trained models construct and update their priors. Our experiments reveal that, although LLMs typically begin with biases that deviate from real-world frequencies, they can approximate Bayesian updating once they see enough in-context evidence. This suggests that the primary limitation in simulating stochastic processes arises from poor priors, not from a failure of ICL itself.

Given these findings, we see both promise and caution for emerging paradigms that treat LLMs as “world models.” In complex domains such as robotics simulations \cite{dagan2023dynamic,song2024trial,zhao2024drivedreamer} or human behavior modeling \cite{aher2023using,park2023generative,moon2024virtual,axtellagent,argyle2023out,loyall1997believable}, accurate simulation relies heavily on well-calibrated base assumptions. Our results underscore that, without calibration or sufficient prompting, LLMs may misrepresent even simple coin-flip dynamics. Yet, once given an adequate stream of observations, these same models exhibit behavior that aligns well with normative Bayesian reasoning.

However, it is worth asking if LLM probabilities ought to be calibrated at all? While we primarily focus on the mechanism in this work, i.e., adjusting LLM probabilities with in-context evidence, we believe that LLMs used as agents should be well-calibrated. One of the primary reasons for this is the growing adoption of LLMs in simulation, particularly probabilistic simulations and world modeling, in which it is quite important to correctly model stochastic outcomes. In addition, well-calibrated models will likely make more fair/unbiased decisions than uncalibrated models \cite{tian2023just}. 

In future work, we would like to explore how our work's discoveries map to multimodal language models. Prior work has shown that vision-language models (VLMs) are blind \cite{rahmanzadehgerviVisionLanguageModels2025} and fail to perform on tasks that are dominated by simple visual reasoning. \citet{petrykALOHaNewMeasure2024} attempted to measure this misalignment in VLMs by analyzing hallucinations in image captioning. Evidently, VLMs fail to accurately correlate visual features with textual prompts, pointing towards hidden miscalibration. Exploring purely visual stochastic tasks and how VLMs perform in those settings is a natural extension to this work.

Overall, this work highlights how ICL can correct miscalibrated priors in a straightforward setting. In more complex scenarios, additional strategies---such as explicit prior calibration or dynamic tuning of prompt design---may be necessary to ensure reliable probabilistic modeling. By grounding our analysis in a simple and interpretable domain, we provide a foundation for further refining the ``LLM-as-world-model'' framework and deepening our understanding of how LLMs handle uncertainty in realistic, evolving environments.

\section{Limitations}
\label{sec:limitations}

While this paper provides insight into how LLMs approximate Bayesian inference in stochastic modeling, our approach has certain limitations that highlight both methodological constraints and fundamental challenges in treating LLMs as Bayesian reasoners.

One key limitation is that our evaluation method captures only a restricted slice of the full posterior distribution. In Bayesian inference, the posterior should account for the entire probability space, but our approach only evaluates the model’s explicit token probabilities for a predefined set of completions. For example, if the expected response is “The coin came up ‘heads’”, the model might alternatively generate “The coin landed on the edge of heads” or “The coin was slightly tilted toward heads”. While we verify that these are low-probability outcomes in our experiments, they still represent probability mass that is not incorporated into our evaluation. If LLMs allocate significant probability to such alternatives, our benchmark may misrepresent their ability to perform Bayesian updates accurately.

Furthermore, while our experiments assess LLM performance in simple Bayesian updating tasks, they do not fully capture the complexities of real-world probabilistic reasoning. Bayesian inference in natural settings often requires reasoning over continuous distributions, hierarchical priors, or distributions with long tails. Our analysis focuses on discrete, categorical predictions, which may not generalize well to more complex probabilistic environments where likelihoods are less structured or where prior distributions must be inferred over high-dimensional latent spaces.

Another methodological limitation arises in evaluating closed-source models. Since our approach relies on extracting logits to approximate posterior distributions, it cannot be directly applied to black-box models such as GPT-4 or Claude. While an alternative approach using sampling could approximate the posterior, this method is costly and susceptible to distortions from API-side interventions such as caching, response smoothing, or temperature adjustments introducing artifacts that obscure the model’s true Bayesian reasoning capabilities.

Beyond these methodological constraints, there are deeper concerns about the limitations of LLMs as Bayesian agents. A fundamental challenge in Bayesian modeling is the specification of a well-calibrated prior. Our findings suggest that LLMs often exhibit poorly calibrated priors when performing in-context learning, which can lead to systematic misestimation in early predictions. While the models do update their beliefs in a manner consistent with Bayesian inference, an inaccurate prior can cause significant initial divergence from the true posterior. This misalignment is particularly concerning in high-stakes applications such as financial forecasting, scientific modeling, and decision-making systems, where incorrect priors can propagate errors through downstream reasoning.

\section*{Acknowledgments}
This paper would not be possible without the Berkeley AI Research (BAIR) espresso machine. It was huddled around this machine did we come up with the idea for this line of questioning in the first place. We would also like to thank Anand Siththaranjan for their review of the paper and providing excellent feedback to make this work better.

As part of their affiliation with UC Berkeley, the authors were supported in part by the National Science Foundation, the Ford Foundation, and/or the Berkeley Artificial Intelligence Research (BAIR) Industrial Alliance program. This material is based upon work supported by the Defense Advanced Research Projects Agency (DARPA), the Army Contracting Command-Aberdeen Proving Grounds (ACC-APG), and the Air Force Research Laboratory under Contract No(s) W912CG-24-C-0011, FA8650-23-C-7316. The views, opinions, and/or findings expressed are those of the authors and should not be interpreted as representing the official views or policies of any supporting entity, including the AFRL, ACC-APG, the Department of Defense or the U.S. Government.

\bibliography{custom}

\appendix

\renewcommand{\theequation}{\thesection.\arabic{equation}}
\renewcommand{\thefigure}{\thesection.\arabic{figure}}
\renewcommand{\thetable}{\thesection.\arabic{table}}

\makeatletter
\@addtoreset{equation}{section}
\@addtoreset{figure}{section}
\@addtoreset{table}{section}
\makeatother

\clearpage
\section*{Appendix}
\label{sec:appendix}

The appendix consists of the following further discussion:
\begin{itemize}
    \item \autoref{app:data} discusses the data used and created in this paper, and the licenses and usage.
    \item \autoref{app:ai} discusses the use of artificial intelligence in the creation of this manuscript.
    \item \autoref{app:methods} explains the methodologies including distribution normalization and comparisons with prior work.
    \item \autoref{app:models} details the models used in this study, their specifications, and training sources.
    \item \autoref{app:multi_coin_flip} presents additional prior results for the coin flipping experiments.
    \item \autoref{app:attentional_correlation} presents additional figures demonstrating the impact of varying $K$, the switchover point.
    \item \autoref{app:multi_die_roll} explores similar results to \autoref{sec:prior} and \autoref{sec:posterior} but with dice rolling (as opposed to coin flips). 
\end{itemize}

\section{Data Usage}
\label{app:data}

This paper relies on several model artifacts including:
\begin{itemize}
    \item Gemma-2 \cite{team2024gemma} released under the \href{https://github.com/google-deepmind/gemma/blob/main/LICENSE}{Gemma license}.
    \item Llama3.1 \cite{dubey2024llama} released under the \href{https://www.llama.com/llama3/license/}{Llama 3 Community License Agreement}.
    \item Phi-3.5 and Phi-3 \cite{abdin2024phi} released under the MIT license.
    \item Mistral 7B \cite{jiang2023mistral} released under the Apache 2.0 license.
    \item Olmo 7B \cite{muennighoff2024olmoe} released under the Apache 2.0 license.
    \item Pythia Scaling Suite \cite{biderman2023pythia} released under the Apache 2.0 license.
\end{itemize}

Our usage of the models is consistent with the above license terms. Our code for computing the analyses in this paper will be released under the MIT license. 

\section{Use of Artificial Intelligence}
\label{app:ai}

This paper includes contributions generated with the assistance of AI tools. Specifically, AI assistants including ChatGPT were used for sentence/paragraph-level editing of the content, the creation of LaTeX tables and figures from raw data sources, and as a coding assistant through GitHub Copilot. All intellectual and creative decisions, including the final content and conclusions, remain the responsibility of the authors. The use of AI in this process was supervised to ensure accuracy and alignment with the intended research outcomes.

\section{Methods}
\label{app:methods}

\subsection{Preliminaries}

We focus on open-source language models, and extract stochastic representations directly from the underlying learned model distributions. For a sequence of tokens, $x = \{x_1,x_2,\dots x_n\}$ in a vocabulary $V$ (of size $|V|$), a large next-token prediction-based language model, $\mathcal{M}$, approximates a probability distribution over the next token: $P_{\mathcal{M}}(x_{i+1}|x_{i},\dots,x_{1})$.  

To evaluate stochastic processes, for each process we define a fixed set of possible ``outcomes'' that a sample from the process can take. Formally, each outcome $o \in \Omega = \{o_1 \dots o_k\}$ is a sequence of tokens corresponding to a string value (for example, when flipping an coin, the outcomes are ``heads'' and ``tails'', corresponding to token sequences \verb|[_heads]| and \verb|[_t,ails]|). For each outcome, we then aim to compute $P_{\mathcal{M}}(o|\text{prompt})$, where the prompt is a sequence of tokens that both (1) describes the process and (2) asks for a sample. While several works estimate this probability by sampling \cite{hopkins2023can,van2024random}, we found that sampling was often unreliable, and thus, we extract this distribution directly from the language model as: \begin{equation}
P_{\mathcal{M}}(o|\text{prompt}) = \prod_{i=1}^{k} P_{\mathcal{M}}(o_i|o_{i-1},\dots,o_1,\text{prompt})
\end{equation}

Note here that for multi-token sequences, we compute the probability conditioned on picking the correct token, and we assume that there is only one unique generator for the sequence $o$. Because these outcomes are a subset of all of the potential token sequences generated by the LLM, we re-normalize the distribution over the support of the options. See \autoref{app:methods_softmax} for more details about the re-normalization process.

In this paper, we primarily measure the total variation distance (TVD) between the true distribution $P^*(o)$ and the normalized model distribution $\hat{P}_{\mathcal{M}}(o)$ over the support $\Omega$: 
\begin{equation}
    \delta(P^*,\hat{P}_{\mathcal{M}}) = \frac{1}{2}\sum_{\omega \in \Omega} \left|P^*(\omega) - \hat{P}_{\mathcal{M}}(\omega)\right|
\end{equation}
The TVD is an intuitive distance measure, which arises as the optimal transport cost between the distributions given a unit cost function. When the TVD is high, the distributions are quite different, and when it is zero, the distributions are identical. 

In this paper, we explore the performance of several models including Gemma-2 \cite{team2024gemma}, Phi-2/Phi-3.5 (mini) \cite{abdin2024phi}, Llama-3.1 (8B) \cite{dubey2024llama}, Mistral 7B \cite{jiang2023mistral} and OLMoE (7B) \cite{muennighoff2024olmoe} along with their instruction-tuned variants. For more details on the models, see \autoref{app:models}.

\subsection{Distribution Normalization}
\label{app:methods_softmax}

Because the set of outcomes $\Omega$ is only a small part of the possible sequences that the LLM can generate, it is often necessary to re-normalize the probability distribution against the support $\Omega$, instead of the full vocabulary space $V$. There are many options that could be picked for re-normalization. In our experiments, we choose to use a linear re-normalization:
\begin{equation}
\hat{P}_{\mathcal{M}}(o) = \frac{P_{\mathcal{M}}(o|\text{prompt})}{\sum_{\omega \in \Omega} P_{\mathcal{M}}(\omega |\text{prompt})}
\end{equation}
This is in contrast to prior work \cite{liu2024llms}, who normalize using a softmax distribution:
\begin{equation}
\hat{P}_{\mathcal{M}}(o) = \frac{exp(P_{\mathcal{M}}(o|\text{prompt}))}{\sum_{\omega \in \Omega} exp(P_{\mathcal{M}}(\omega |\text{prompt}))}
\end{equation}
Unfortunately, in the limit of small probabilities, for $p_i, 1 < i < |\Omega|$, as $p_i \to 0$:
\begin{equation}
    \lim_{p_i \to 0, p_j \to 0} \frac{e^{p_i}}{\sum_je^{p_j}} = \frac{1}{\sum_je^{p_j}} \approx \frac{1}{|\Omega|}
\end{equation}
This can significantly impact the computation of downstream measures. Normalizing linearly avoids this issue, but can sometimes cause numeric instability. 

\subsection{Instruct Models Chat Templates}

In order to make instruction-tuned models compatible with our formulation for extracting token probabilities, we employ chat templates in the following manner. 
First, we construct the chat as follows: 
\begin{lstlisting}[language=Python]
    chat = {'user': instruct_prompt, 'assistant': input_prompt}
\end{lstlisting}
Here \texttt{instruct\_prompt} explicitly provides a directive to the LLM such as \texttt{"Please toss a coin and tell me whether it landed on heads or tails."}. 
The \texttt{input\_prompt} field contains the portion to be completed by the LLM spoken in the first person, such as \texttt{"I tossed a coin and it landed on"}. 
Next, in order to properly extract token probabilities pertaining to the support of the random process at the end of the input sequence, we modify models' chat templates to omit the EOT tokens. 

\subsection{Prompt Normalization}

The specific prompt itself may have an impact on the outcomes of the experiment, given the myriad of ways that a coin flip can be specified. To help alleviate this variance, in all experiments in the paper we normalize across a large set of 50 potential result prompts, shown in \autoref{fig:prompts}, and 5 instruct prompts, shown in \autoref{fig:instruct_prompts}.

\begin{figure}
    \centering
    \begin{lstlisting}
I just flipped a coin, and it came up 
The result of my coin flip is 
After flipping the coin, it landed on 
I flipped a coin, and the outcome was 
The coin flip resulted in 
After tossing the coin, it ended up on 
I tossed a coin, and it fell on 
The coin I flipped landed on 
The coin I tossed resulted in 
Following the coin flip, it showed 
The coin fell on 
The flip of the coin resulted in 
I flipped the coin, and it settled on 
The result after flipping the coin is 
The outcome of my coin flip is 
I tossed the coin, and the outcome is 
The result of my coin toss is 
I flipped the coin, and it came up 
The coin came to rest on 
After flipping, the coin showed 
The toss of the coin revealed 
I flipped the coin, and it turned up 
The coin toss ended with 
After tossing the coin, it showed 
The coin flipped over to 
After flipping, the coin settled on 
My coin toss resulted in 
The outcome of my coin flip turned out to be 
I flipped the coin, and its final position was 
The coin fell, showing 
I tossed the coin, and it landed showing 
Following the toss, the coin showed 
The flip resulted in the coin landing on 
The coin toss revealed 
The outcome of the coin landing is 
After tossing, the coin landed on 
I flipped the coin and saw it land on 
After the flip, the coin showed 
The result of tossing the coin was 
When I flipped the coin, it landed on 
The coin showed this side after the flip: 
The flip of the coin ended with 
After tossing, the coin fell to show 
The result of my toss came out as 
The toss of the coin came to rest on 
The coin after the flip landed on 
I flipped the coin, and it ended on 
The result of the coin toss ended up being 
I flipped a coin, and its final side was 
The coin flip showed the result: 
    \end{lstlisting}
    \caption{A list of possible prompts describing a coin flip result.}
    \label{fig:prompts}
\end{figure}

\begin{figure}
    \centering
    \begin{lstlisting}
Please complete this sentence: I just flipped a coin, and it landed on
Finish this sentence: The result of my coin flip is
Complete the sentence: After flipping the coin, it landed on
Fill in the rest: I flipped a coin, and the outcome was
Complete the phrase: The coin flip resulted in
    \end{lstlisting}
    \caption{A list of possible instruct prompts describing a coin flip result.}
    \label{fig:instruct_prompts}
\end{figure}

\section{Models}
\label{app:models}

\paragraph{google/gemma-2-2b}: Gemma-2 \cite{team2024gemma} is a 2-billion parameter model from Google DeepMind, built on a decoder-only Transformer architecture. It uses a model dimension of 2304, 26 layers, 8 heads with Grouped-Query Attention (GQA), and a feedforward dimension of 18,432. The model features 256,128 vocabulary entries using a SentencePiece tokenizer optimized for multiple languages. Trained on 2 trillion tokens from sources like web documents, code, and scientific articles, it incorporates local sliding window attention (4096 tokens) and global attention (8192 tokens). It also uses knowledge distillation from larger models, which enables it to achieve strong results with more efficient training.

\paragraph{google/gemma-2-2b-it:} Gemma-2-IT \cite{team2024gemma} is the instruction-tuned version of Gemma-2, featuring the same architecture as Gemma-2-2B. The model has been fine-tuned on supervised datasets and further improved using RLHF (Reinforcement Learning from Human Feedback) for better instruction-following capabilities. It uses the same 256,128-entry vocabulary and was trained on similar data sources. Gemma-2-IT includes additional tuning to enhance safety and reduce hallucinations.

\paragraph{meta-llama/llama-3.1-8B:} Llama-3 \cite{dubey2024llama} is a foundation model developed by Meta, built with an 8 billion parameter dense Transformer architecture. The model has 32 layers, a model dimension of 4096, a feedforward dimension of 14,336, and 32 attention heads. It supports multilingual tasks, coding, and reasoning with a context window of 8K tokens. Llama-3 was pre-trained on a dataset of 15 trillion tokens, spanning a variety of sources such as web documents, code, and multilingual texts, with a vocabulary size of 128,000 tokens using a tokenizer optimized for multilingual use.

\paragraph{meta-llama/llama-3.1-8B-Instruct:} Llama-3-Instruct \cite{dubey2024llama} is the instruction-tuned variant of Llama-3, also comprising 8 billion parameters, 32 layers, 4096 model dimensions, and a feedforward dimension of 14,336. This version is fine-tuned to follow human instructions better, leveraging supervised fine-tuning and Direct Preference Optimization (DPO). It is designed for tasks requiring precise instruction following, including coding, reasoning, and complex dialogue, while supporting tools like code generation and multilingual text processing. It also includes additional tuning to enhance safety and reduce hallucinations.

\paragraph{microsoft/phi-3.5-mini-instruct:} Phi-3 \cite{abdin2024phi} is a 3.8-billion parameter Transformer model designed by Microsoft, optimized for both small-scale deployment and high-performance tasks. The model has 32 layers, 3072 hidden dimensions, 32 attention heads, and a default context length of 4K tokens, extendable to 128K using LongRope. It was trained on 3.3 trillion tokens, with a dataset comprising heavily filtered publicly available web data and synthetic data. Its instruction-following capability is enhanced through supervised fine-tuning and Reinforcement Learning from Human Feedback (RLHF)

\paragraph{microsoft/phi-2:} Phi-2 \cite{abdin2024phi} is a 2.7-billion parameter model, part of Microsoft's Phi series, designed for efficient performance in smaller-scale models. Like Phi-3, it uses a transformer-based decoder architecture with Grouped-Query Attention (GQA) and a vocabulary size of 320641 tokens and is trained on a mixture of filtered web data and LLM-generated synthetic data.

\paragraph{mistalai/Mistral-7B:} Mistral-7B \cite{jiang2023mistral} is a 7-billion parameter model developed by Mistral AI, built with a Transformer architecture optimized for efficiency and performance. The model has 32 layers, a model dimension of 4096, a feedforward dimension of 14,336, and 32 attention heads. Mistral-7B uses  Grouped-Query Attention (GQA) and Sliding Window Attention (SWA) to handle sequences up to 8192 tokens.

\paragraph{mistralai/Mistral-7B-Instruct:} Mistral-7B-Instruct  \cite{jiang2023mistral} is the instruction-tuned variant of Mistral-7B, featuring the same architecture with 7 billion parameters, 32 layers, 4096 model dimensions, and a feedforward dimension of 14,336.

\paragraph{allenai/OLMoE-1B-7B:} OLMoE-1B-7B \cite{muennighoff2024olmoe} is a Mixture-of-Experts LLM with 1B active and 7B total parameters developed by Allen AI, designed for open access and transparency. The model consists of 32 layers, a model dimension of 4096, a feedforward dimension of 11,008 (due to its SwiGLU activation), and 32 attention heads. The vocabulary size is 50,280 tokens, based on a modified BPE tokenizer that includes special tokens for anonymizing personally identifiable information (PII). OLMo-7B was trained on Dolma, which comprises 2.46 trillion tokens from diverse sources like Common Crawl, GitHub, Wikipedia, and scientific papers. 

\paragraph{allenai/OLMoE-1B-7B-Instruct:} OLMoE-1B-7B-Instruct \cite{muennighoff2024olmoe} is a Mixture-of-Experts LLM with 1B active and 7B total parameters that has been adapted via SFT and DPO from OLMoE-1B-7B. Like OLMoE-1B-7B, it features 32 layers, a model dimension of 4096, and 32 attention heads, with a feedforward dimension of 11,008. This variant was fine-tuned using a mixture of human-annotated and distilled instruction data, optimized further using Direct Preference Optimization (DPO) for better alignment with human preferences. 

\paragraph{Pythia Scaling Suite:} Pythia \cite{biderman2023pythia} is a suite of 16 publicly available autoregressive language models, spanning parameter sizes from 70M to 12B, designed to facilitate scientific research into the dynamics of training and scaling in large language models. Each model in the suite was trained on the Pile dataset in a controlled, consistent manner, ensuring identical data ordering and architecture across scales. The suite includes models trained on both the original Pile dataset and a deduplicated version to allow comparative studies of data redundancy effects. Pythia’s intermediate checkpointing—offering 154 checkpoints per model—enables detailed longitudinal studies of model behavior over training. 

\section{Additional Results}
\label{app:multi_coin_flip}

In this section, we present additional results for the coin flip experiments in \autoref{sec:prior} and \autoref{sec:posterior}.

\subsection{Longer Convergence Chains}

\begin{figure*}
    \centering
    \small
    \includegraphics[width=\linewidth]{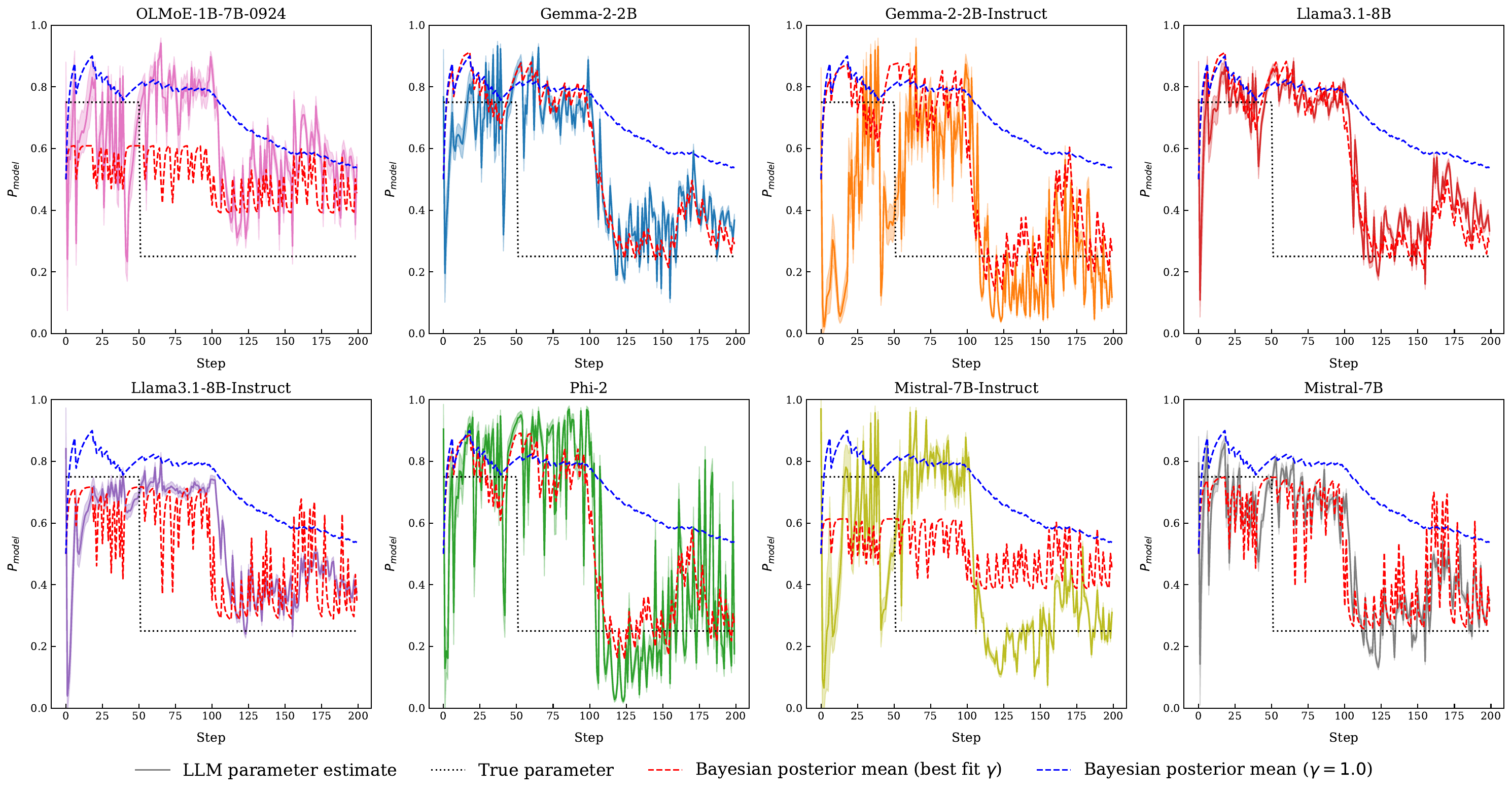}
    \caption{\textbf{Posterior evolution during Bayesian filtering:} The figure shows a single rollout of classical Bayesian filtering alongside model predictive probabilities in a 200-sample coin flip ICL task.}
    \label{fig:posterior_chain_200}
\end{figure*}

In addition to a roll-out of length 100, we also looked at a roll-out of length 200, with the trajectory given in \autoref{fig:posterior_chain_200}. We can see that in general, the convergence pattern matches the $100$ sample case. 

\begin{figure}[t]
    \centering
    \includegraphics[width=\linewidth]{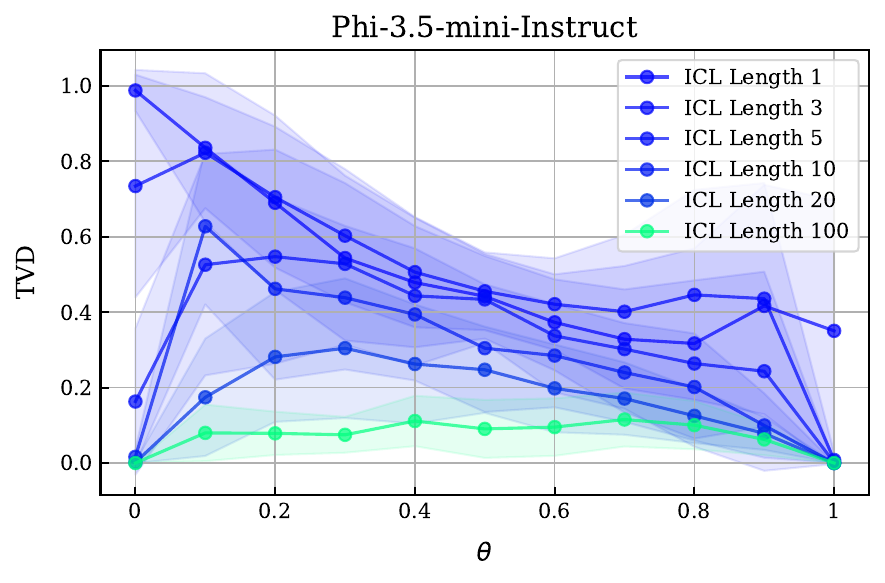}
    \caption{Mean total variation distance (TVD, $\downarrow$) vs. bias percentage for several ICL example lengths on the coin flipping task for the Phi-3.5-mini-instruct model.}
    \label{fig:coin_flip_icl_phi_3_5_mini_instruct}
\end{figure}

\begin{figure}[t]
    \centering
    \includegraphics[width=\linewidth]{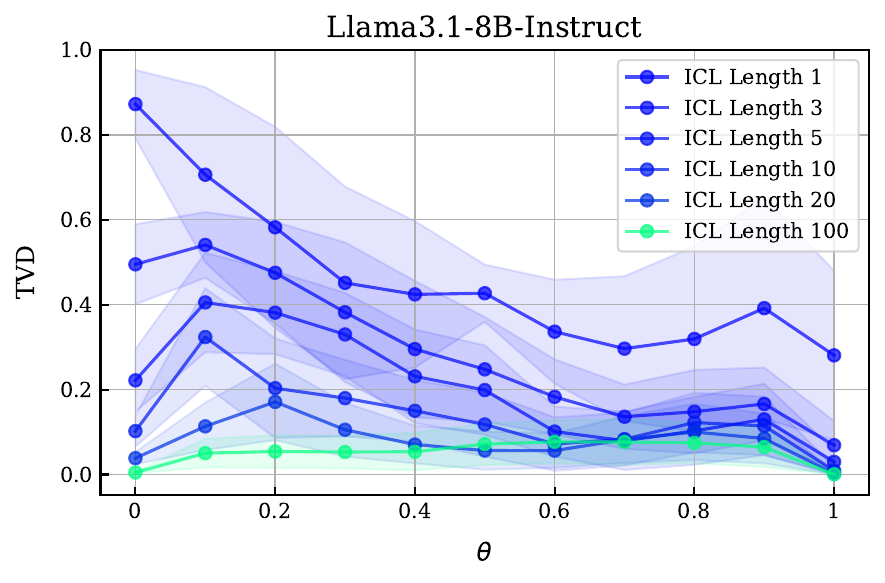}
    \caption{Mean total variation distance (TVD, $\downarrow$) vs. bias percentage for several ICL example lengths on the coin flipping task for the Llama-3.1-8B-Instruct model.}
    \label{fig:coin_flip_icl_llama_3_1_8B_instruct}
\end{figure}

\begin{figure}[t]
    \centering
    \includegraphics[width=\linewidth]{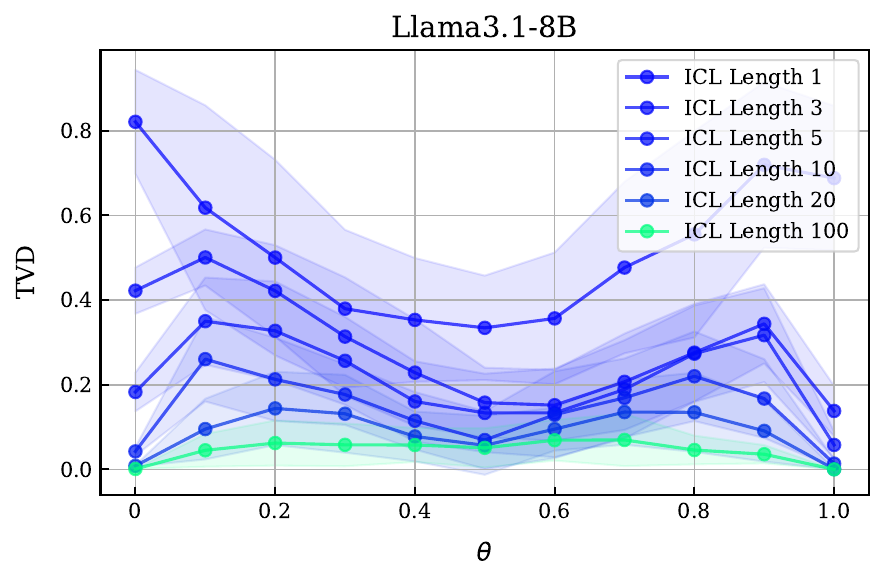}
    \caption{Mean total variation distance (TVD, $\downarrow$) vs. bias percentage for several ICL example lengths on the coin flipping task for the Llama-3.1-8B model.}
    \label{fig:coin_flip_icl_llama_3_1_8B}
\end{figure}

\begin{figure}[t]
    \centering
    \includegraphics[width=\linewidth]{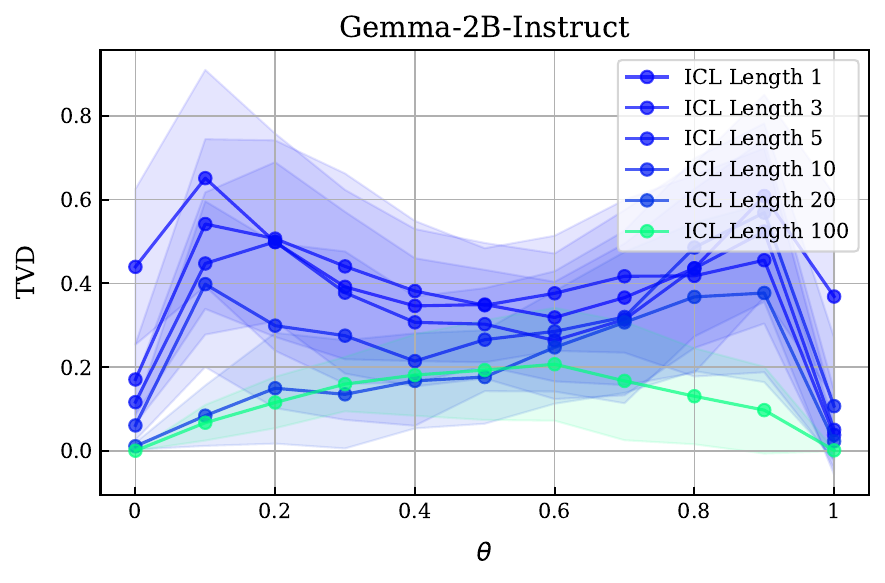}
    \caption{Mean total variation distance (TVD, $\downarrow$) vs. bias percentage for several ICL example lengths on the coin flipping task for the Gemma-2-2B-IT model.}
    \label{fig:coin_flip_icl_gemma_2_2b_it}
\end{figure}

\begin{figure}[t]
    \centering
    \includegraphics[width=\linewidth]{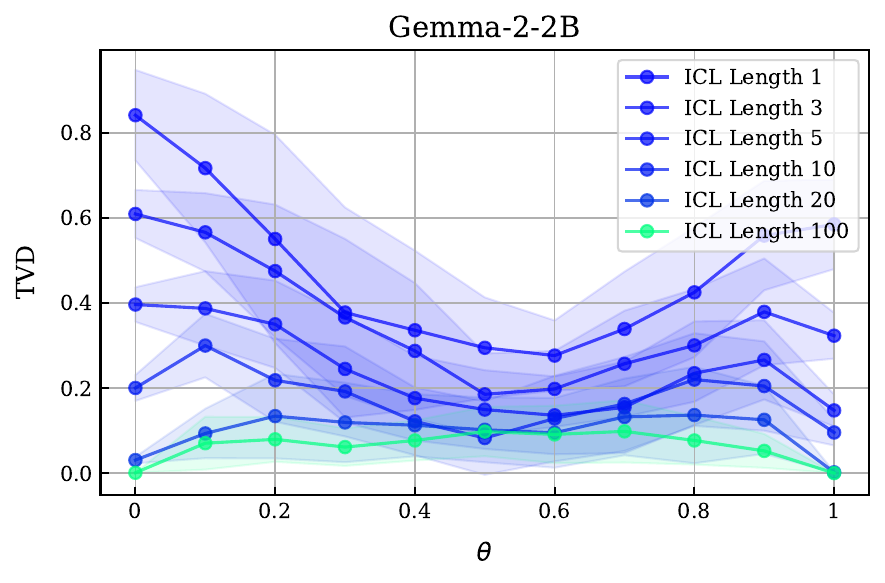}
    \caption{Mean total variation distance (TVD, $\downarrow$) vs. bias percentage for several ICL example lengths on the coin flipping task for the Gemma-2-2B model.}
    \label{fig:coin_flip_icl_gemma_2_2b}
\end{figure}

\subsection{ICL Scaling Results}
\label{app:icl_scaling}

In \autoref{fig:coin_flip_icl_phi_3_5_mini_instruct}, \autoref{fig:coin_flip_icl_llama_3_1_8B_instruct}, \autoref{fig:coin_flip_icl_llama_3_1_8B}, \autoref{fig:coin_flip_icl_gemma_2_2b_it}, and \autoref{fig:coin_flip_icl_gemma_2_2b}, we present the Mean total variation distance (TVD, $\downarrow$) against bias percentage for several ICL (In-Context Learning) example lengths across different models. These plots help analyze how well each model handles bias in a coin flip prediction task as the ICL context varies. The lower the TVD score, the better the model performs in generating unbiased predictions.

\begin{figure*}
    \centering
    \includegraphics[width=0.33\linewidth]{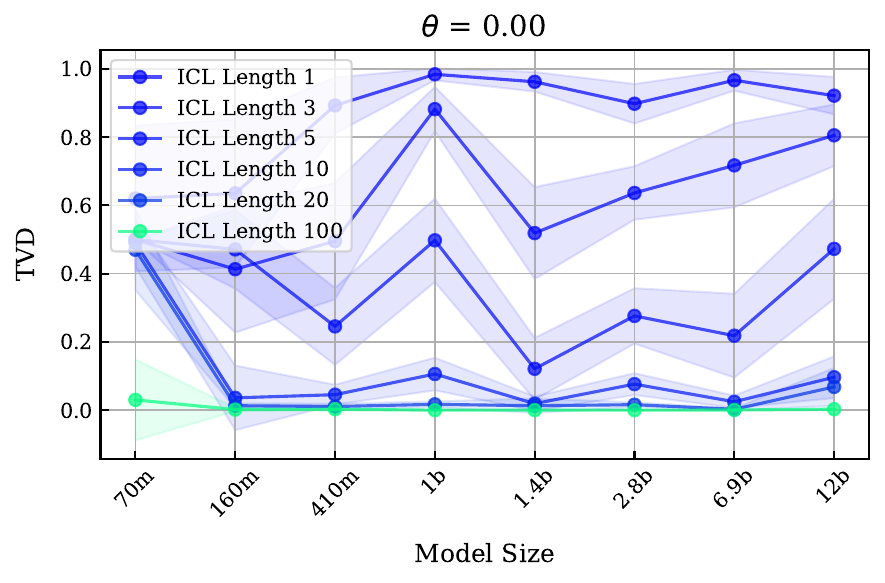}
    \includegraphics[width=0.33\linewidth]{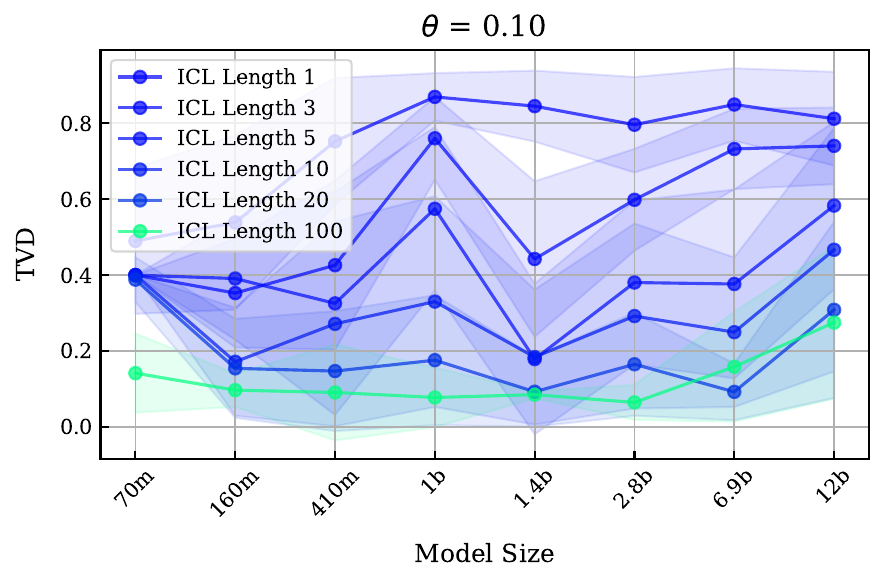}
    \includegraphics[width=0.33\linewidth]{figures/pythia_icl_tvd_20.pdf}
    \includegraphics[width=0.33\linewidth]{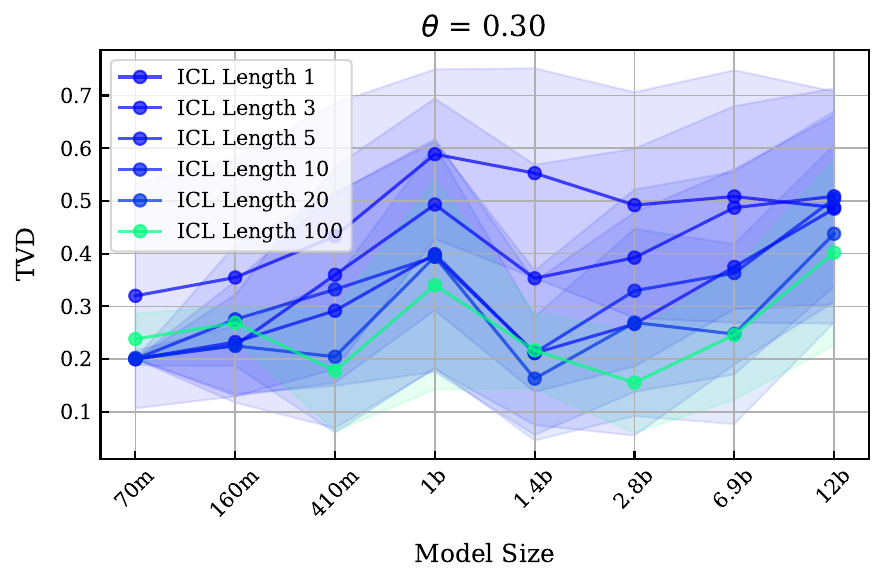}
    \includegraphics[width=0.33\linewidth]{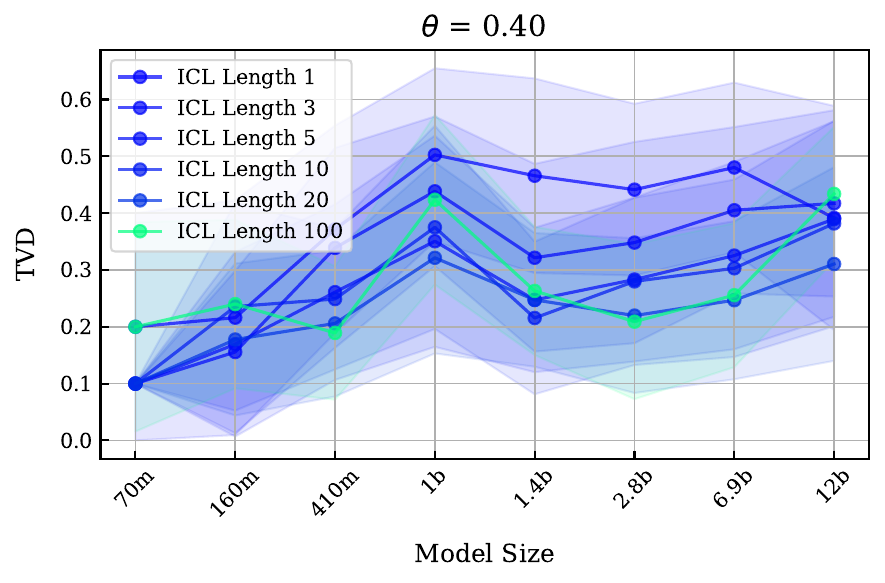}
    \includegraphics[width=0.33\linewidth]{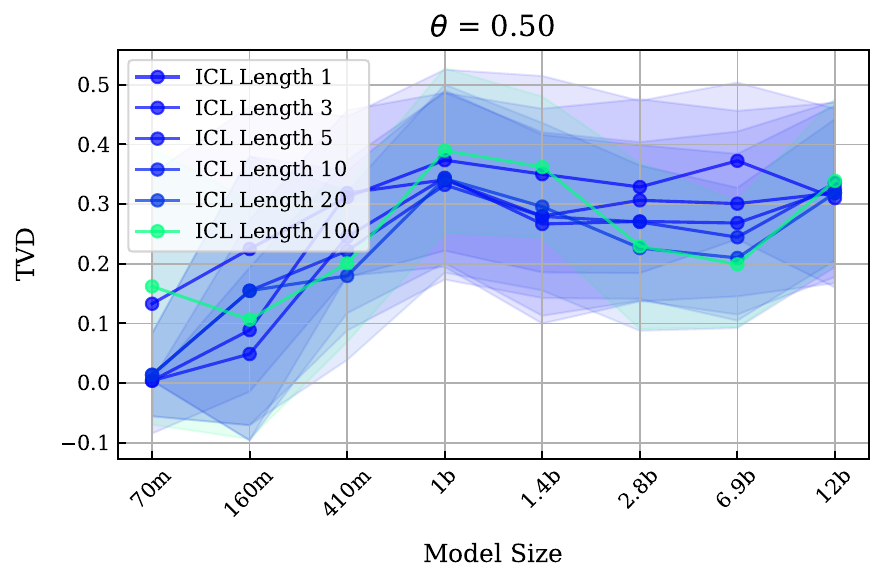}
    \includegraphics[width=0.33\linewidth]{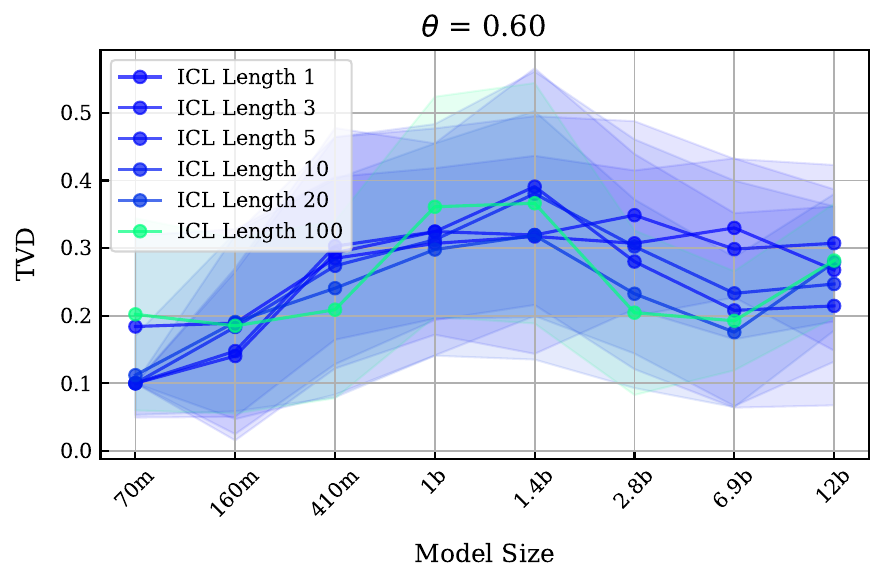}
    \includegraphics[width=0.33\linewidth]{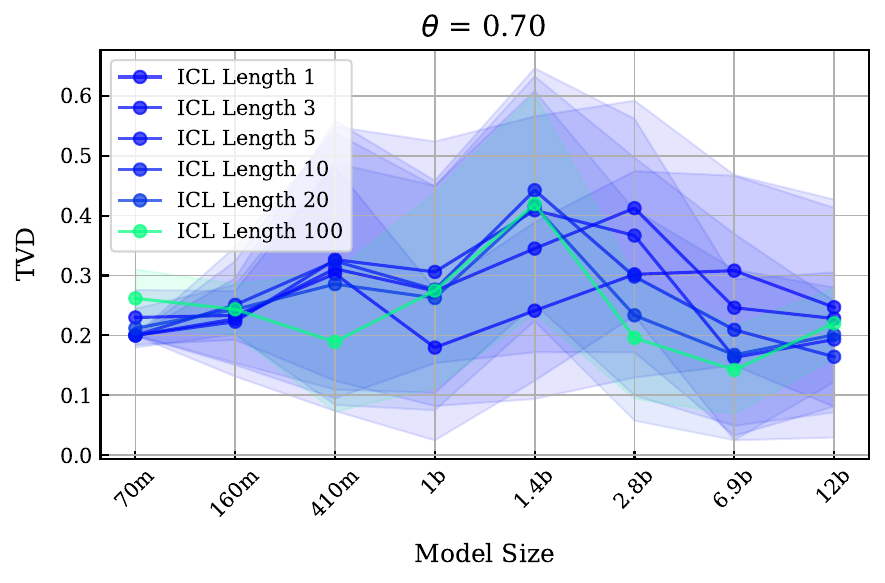}
    \includegraphics[width=0.33\linewidth]{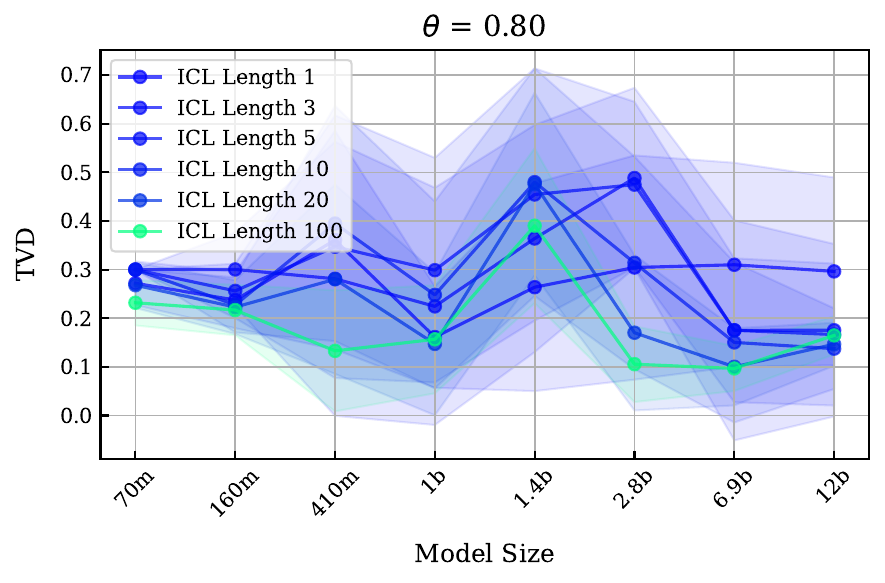}
    \includegraphics[width=0.33\linewidth]{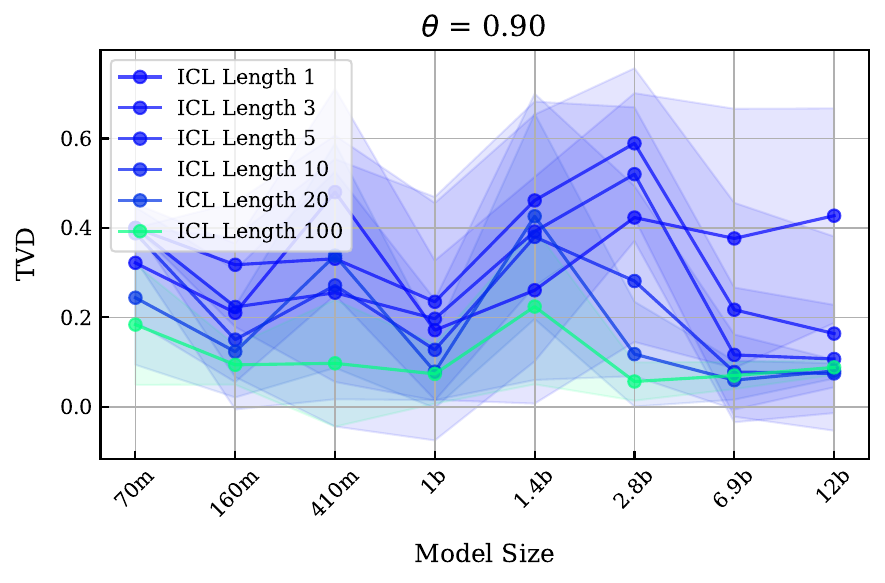}
    \includegraphics[width=0.33\linewidth]{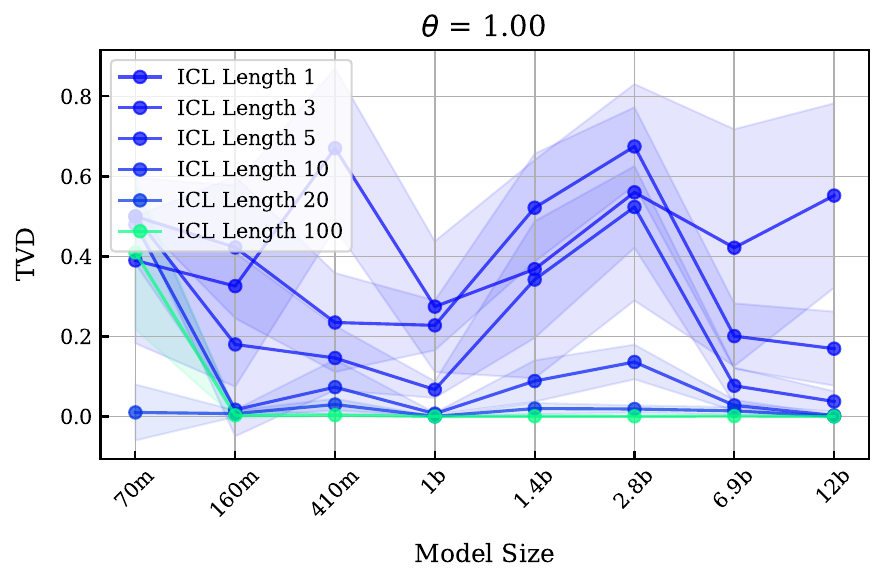}
    \caption{\textbf{ICL and parameter scaling:} Mean total variation distance (TVD, $\downarrow$) vs. model size across the Pythia Scaling Suite family with a biasing statement for all values of $\theta$. Model size does not have a clear impact on the benefits from ICL.}
    \label{fig:app_icl_scaling}
\end{figure*}

In \autoref{fig:app_icl_scaling}, we present all the results from the ICL scaling experiments in Section \ref{sec:icl_scaling}.

\section{Varying the Switchover Point}
\label{app:attentional_correlation}

\begin{figure}
    \centering
    \includegraphics[trim={3mm 0 0 0},clip,width=\linewidth]{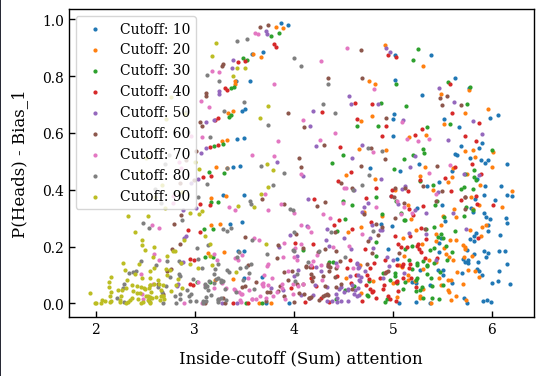}
    \caption{This plot shows the correlation between the amount of attention paid within the cutoff region and the calibration accuracy.}
    \label{fig:app_att_1}
\end{figure}

\begin{figure}
    \centering
    \includegraphics[width=\linewidth]{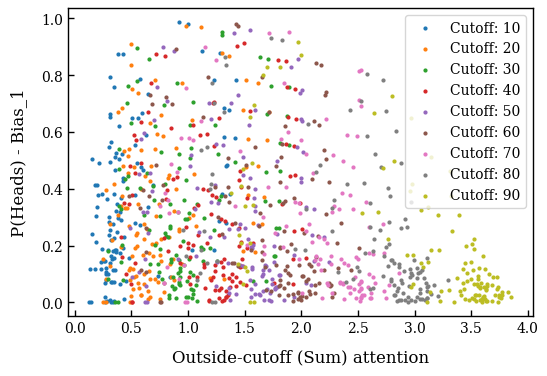}
    \caption{This plot shows the correlation between the amount of attention paid outside the cutoff region and the calibration accuracy.}
    \label{fig:app_att_2}
\end{figure}

We also perform several experiments varying the value $K$ (the switchover point) in the  experiments. \autoref{fig:app_att_1} shows the correlation between the amount of attention paid within the cutoff region and the calibration accuracy, where we see that while the size of the cutoff ($K$) does impact the amount of attention paid to the model, there is little correlation between that amount of attention and the calibration accuracy. Similarly, \autoref{fig:app_att_2} shows the correlation between the amount of attention paid outside the cutoff region and the calibration accuracy, demonstrating a similar lack of correlation.

\begin{figure}
    \centering
    \includegraphics[width=\linewidth]{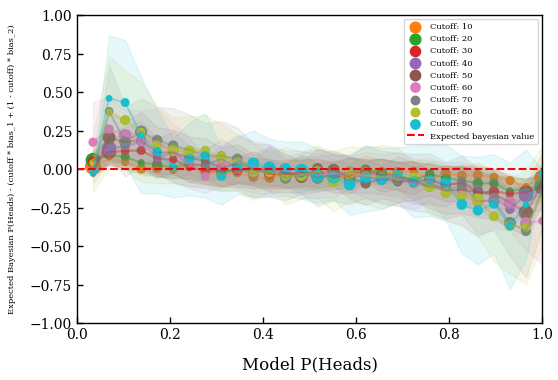}
    \caption{Plot showing the deviation of the model predicted $\theta$ against the expected Bayesian update probability for different values of $K$.}
    \label{fig:app_att_3}
\end{figure}

These results are further shown in \autoref{fig:app_att_3} which plots the deviation of the value $\theta$ against the expected Bayesian update probability for different values of $K$. We can see that as the probabilities become more extreme, the deviation becomes higher, and models have more trouble adjusting to more extreme probabilities, however there is no statistically significant difference between the $K$ values. 

\section{Rolling Dice}
\label{app:multi_die_roll}

To explore the applicability of our results beyond coin flips, we also experiment with a similar simple distribution, rolling dice.  We then ask the LLM to complete the prompt  \texttt{``I rolled a die and it landed on''} over the choices of one through six. For biased variants, we provided explicit biasing statements within prompts to the model such as: ``When I flip coins, they land on heads X\% of the time,'' where X is a percentage between 0\% and 100\%, or ``When I roll dice, they land on N X\% of the time.'' 

\begin{figure*}
    \centering
    \hspace{\stretch{1}}
    \includegraphics[width=0.49\linewidth]{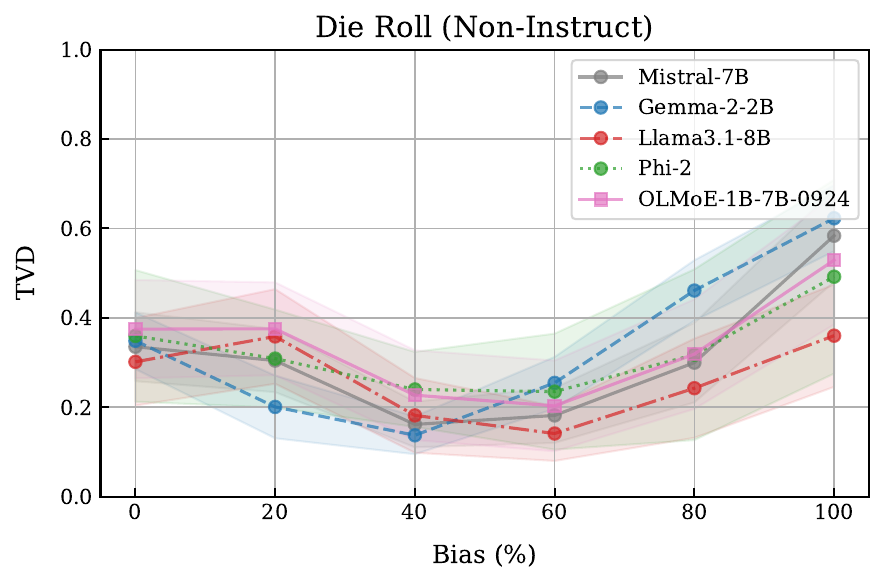}
    \hspace{\stretch{1}}
    \includegraphics[width=0.49\linewidth]{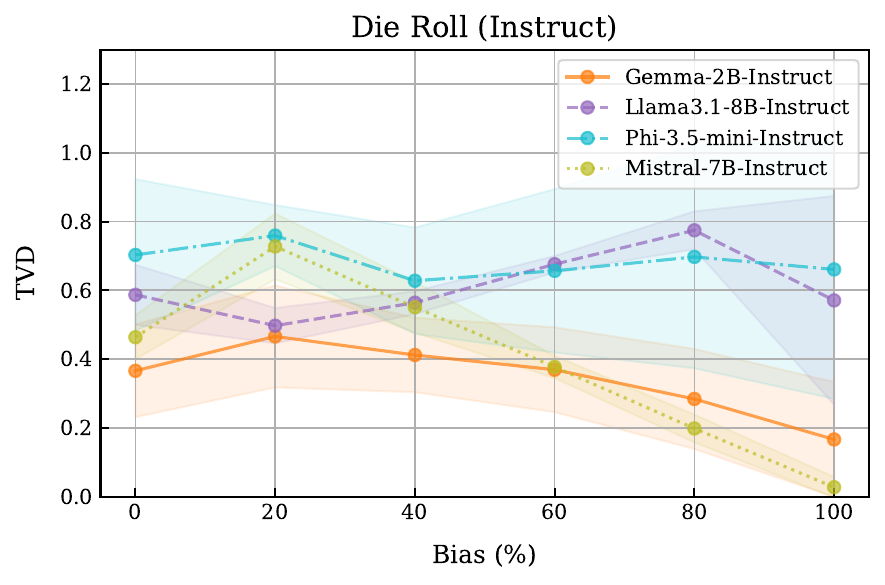}
    \hspace{\stretch{1}}
    \caption{\textbf{Biased die rolls:} Plots of mean total variation distance (TVD, $\downarrow$) against bias percentage for non-instruct (left) and instruct (right) models when aggregated across prompts (N=50) for the biased die rolling experiment.}
    \label{fig:die_roll}
\end{figure*}

The results are shown in \autoref{fig:die_roll}. For each bias percentage, we averaged results across the six die faces and 50 prompt variants, totaling 300 trials per bias percentage. Non-instruct models generally performed better than their instruct counterparts, and best around a 50\%-60\% bias, struggling more with higher biases. Instruct model performance was more varied, with some models showing little change in behavior and others improving as the bias value increased.

Results on die-rolling for in-context learning are shown below. While both instruction finetuned and non-instruction-finetuned variants benefit from increasing numbers of examples, the non-instruction-finetuned variants benefit more and generally exhibit better performance.

\begin{figure}
    \centering
    \includegraphics[width=\linewidth]{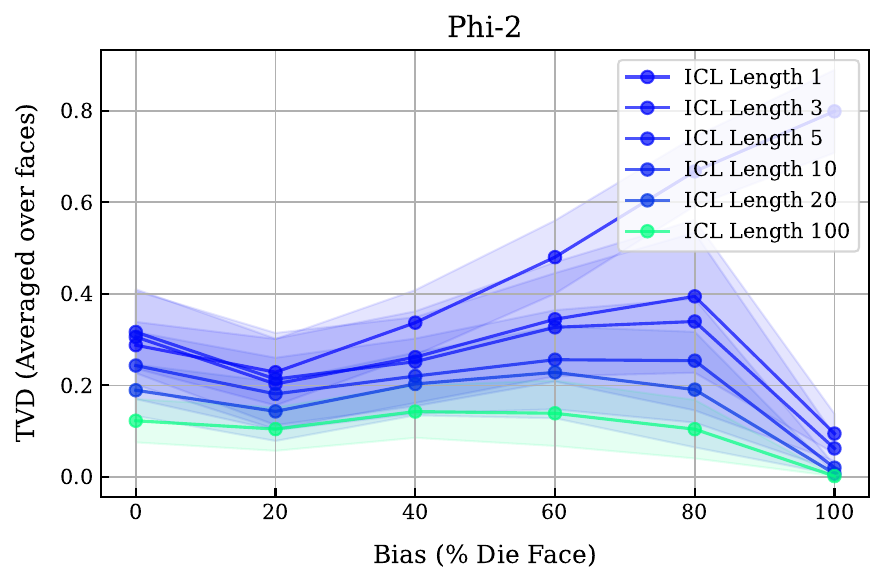}
    \caption{Mean total variation distance (TVD, $\downarrow$) vs. bias percentage for several ICL example lengths on the die rolling task for the Microsoft Phi-2 model.}
    \label{fig:phi_2_die_roll}
\end{figure}

\begin{figure}
    \centering
    \includegraphics[width=\linewidth]{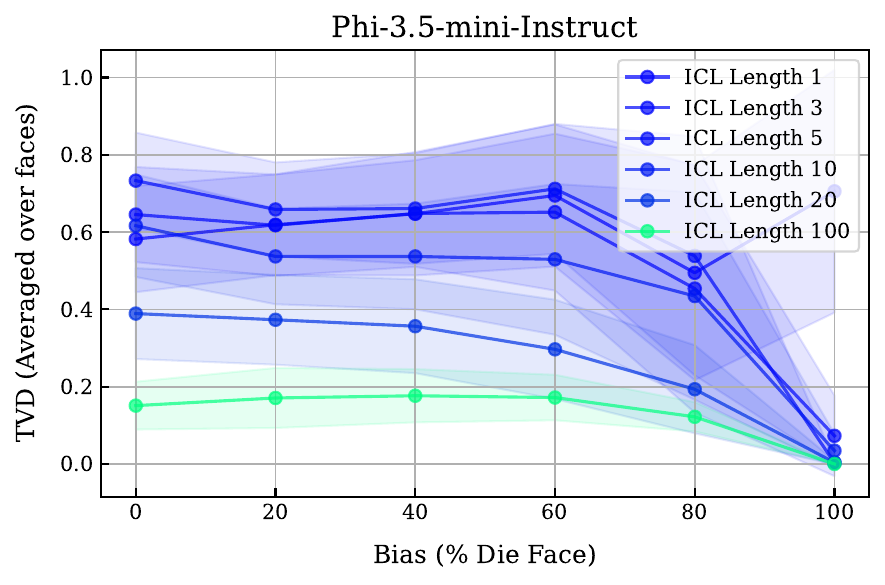}
    \caption{Mean total variation distance (TVD, $\downarrow$) vs. bias percentage for several ICL example lengths on the die rolling task for the Microsoft Phi-3.5-mini-instruct model.}
    \label{fig:phi_3_5_mini_instruct_die_roll}
\end{figure}

\begin{figure}
    \centering
    \includegraphics[width=\linewidth]{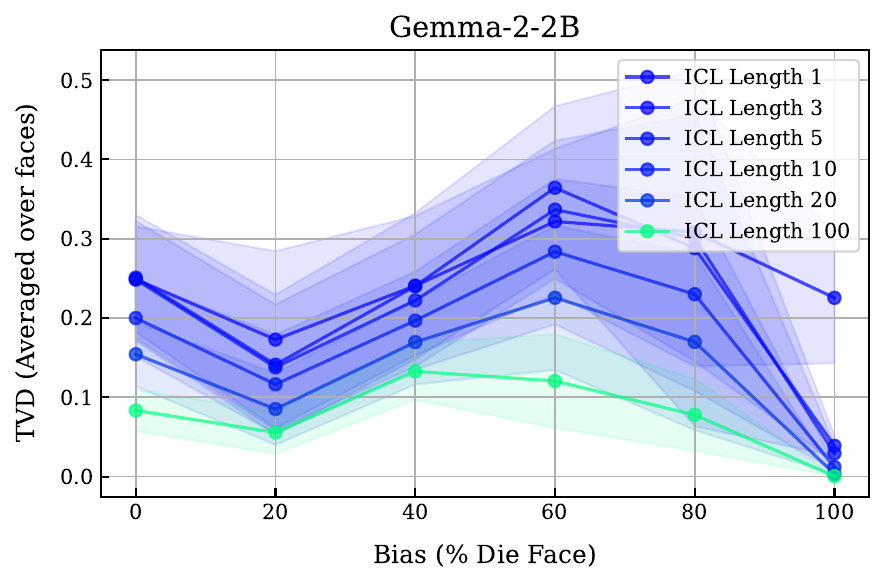}
    \caption{Mean total variation distance (TVD, $\downarrow$) vs. bias percentage for several ICL example lengths on the die rolling task for the Google Gemma-2-2B model.}
    \label{fig:gemma_2_2b_die_roll}
\end{figure}

\begin{figure}
    \centering
    \includegraphics[width=\linewidth]{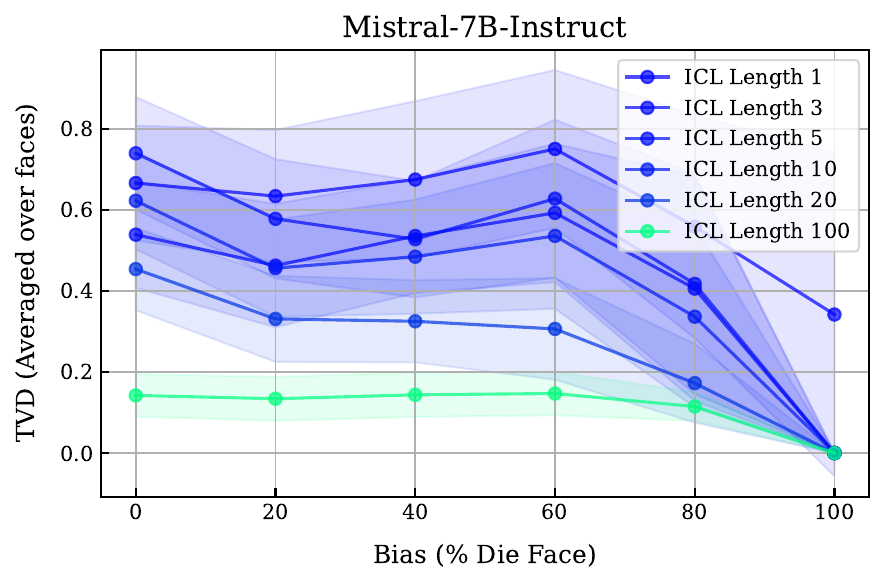}
    \caption{Mean total variation distance (TVD, $\downarrow$) vs. bias percentage for several ICL example lengths on the die rolling task for the Mistral-7B-Instruct model.}
    \label{fig:mistral_7b_instruct_die_roll}
\end{figure}

\begin{figure}
    \centering
    \includegraphics[width=\linewidth]{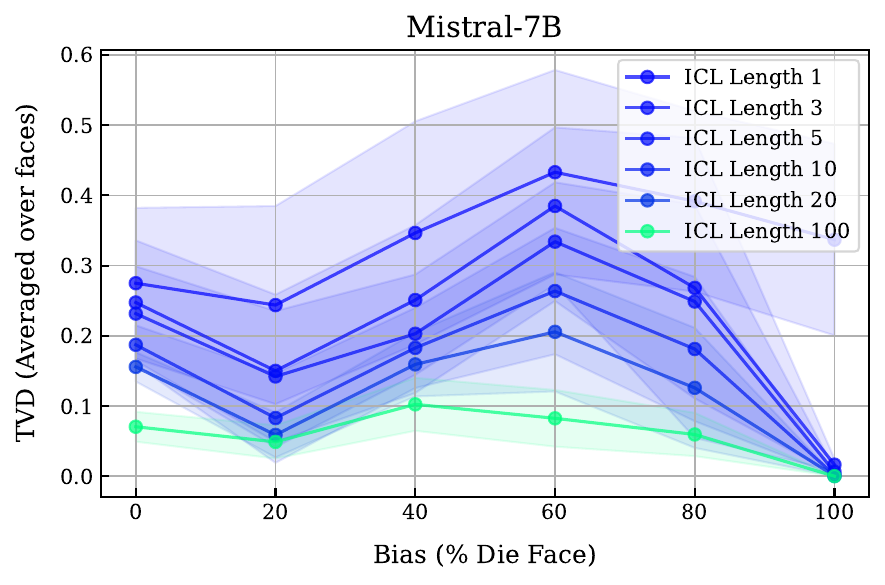}
    \caption{Mean total variation distance (TVD, $\downarrow$) vs. bias percentage for several ICL example lengths on the die rolling task for the Mistral-7B model.}
    \label{fig:mistral_7b_die_roll}
\end{figure}

In \autoref{fig:phi_2_die_roll}, \autoref{fig:phi_3_5_mini_instruct_die_roll}, \autoref{fig:gemma_2_2b_die_roll}, \autoref{fig:mistral_7b_instruct_die_roll}, and \autoref{fig:mistral_7b_die_roll}, we present ICL plots measuring TVD for a variety of model variants on the simple dice rolling experiment. These results correlate well with the results observed in  \autoref{sec:prior}, the coin flip experiments. 

\begin{figure}
    \centering
    \includegraphics[width=\linewidth]{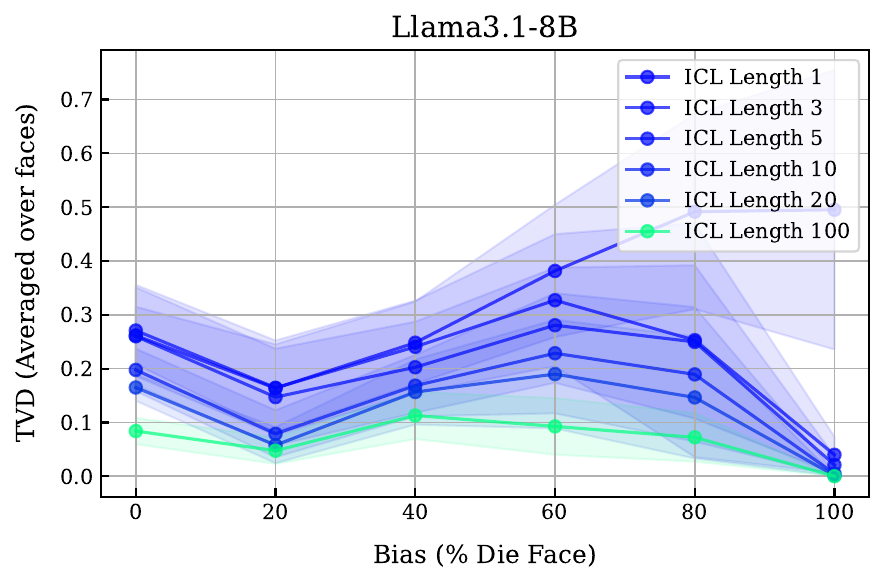}
    \caption{Mean total variation distance (TVD, $\downarrow$) vs. bias percentage for several ICL example lengths on the die rolling task for the Llama3.1-8B model.}
    \label{fig:llama3.1_die_roll}
\end{figure}

\begin{figure}
    \centering
    \includegraphics[width=\linewidth]{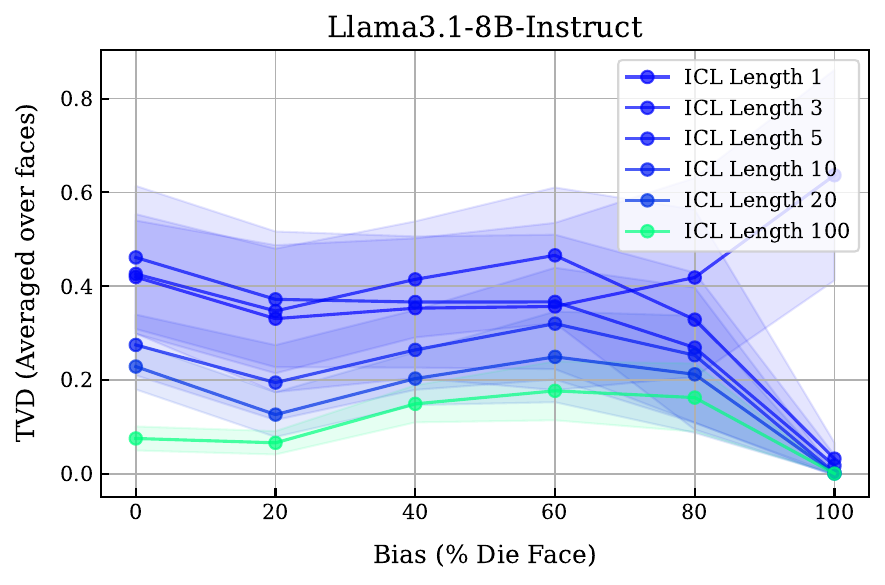}
    \caption{Mean total variation distance (TVD, $\downarrow$) vs. bias percentage for several ICL example lengths on the die rolling task for the Llama3.1-8B-Instruct model.}
    \label{fig:llama3.1_instruct_die_roll}
\end{figure}

\end{document}